\documentclass{article} 

\usepackage{makecell}
\usepackage{caption}
\usepackage{bicaption}
\usepackage{multirow}
\usepackage{graphicx}
\usepackage{float}
\usepackage{hyperref}

\usepackage{hyperref, tikz, float, subfigure, multicol, amsmath, amsthm, alphalph, amsfonts, amssymb, geometry, enumitem, parskip, xcolor, sectsty}
\usepackage[normalem]{ulem}
\usepackage[font=scriptsize,labelfont=bf]{caption}
\usepackage[explicit]{titlesec}
\usepackage[scaled]{helvet}
 
\usepackage[T1]{fontenc}

\usepackage{setspace}
\usepackage{CJKutf8}
\usepackage{booktabs}

\usepackage{fancyhdr}
\definecolor{ghent_blue}{rgb}{0.1176, 0.392, 0.7843}
\definecolor{ghent_dark}{rgb}{0.0, 0.2, 0.4}
\title{{\color{ghent_blue} TITLE: The comparison of translationese in machine translation and human transation in terms of translation relations}}

\date{}         
\sectionfont{\fontsize{16}{15}\selectfont}
\sectionfont{\color{ghent_blue}}

\hypersetup{colorlinks=true,linkcolor=black}
\begin{document}



\thispagestyle{empty}
\vspace*{1cm}
\fontsize{20}{30}\selectfont{\color{ghent_blue}\hfill\break\textbf{{THE COMPARISON BETWEEN MACHINE TRANSLATION AND HUMAN TRANSLATION IN TERMS OF TRANSLATION RELATIONS  
}}}\\
\fontsize{15}{30}\selectfont{\color{ghent_blue}\hfill\break\textbf{
{MASTER'S THESIS AT GHENT UNIVERSITY AND KU LEUVEN 
}}}
\fontsize{15}{30}\selectfont{\color{ghent_blue}\hfill\break\textbf{
{Faculty of Arts and Philosophy 
}}}\\

\noindent\fontsize{15}{0}\selectfont Fan Zhou
\vspace{3mm}\\
\fontsize{10}{0}\selectfont Student number: 02016257  \vspace{1cm}\\
\noindent\fontsize{12}{0}\selectfont Supervisor(s): Dr. Arda Tezcan\\ 
\fontsize{12}{0}\selectfont Co-supervisor(s): Prof. Linda Badan \vspace{1cm}\\ 
\noindent\fontsize{10}{0}\selectfont A thesis submitted to Ghent University in partial fulfilment of the requirements for the degree of Master of Arts in Linguistics: Natural Language Processing \vspace{1cm}\\
\noindent Ghent University and KU Leuven\\
Academic year: 2021 - 2022 
\vspace*{\fill}
\begin{flushleft}
\begin{figure}[b!]
\includegraphics[scale = 1.4]{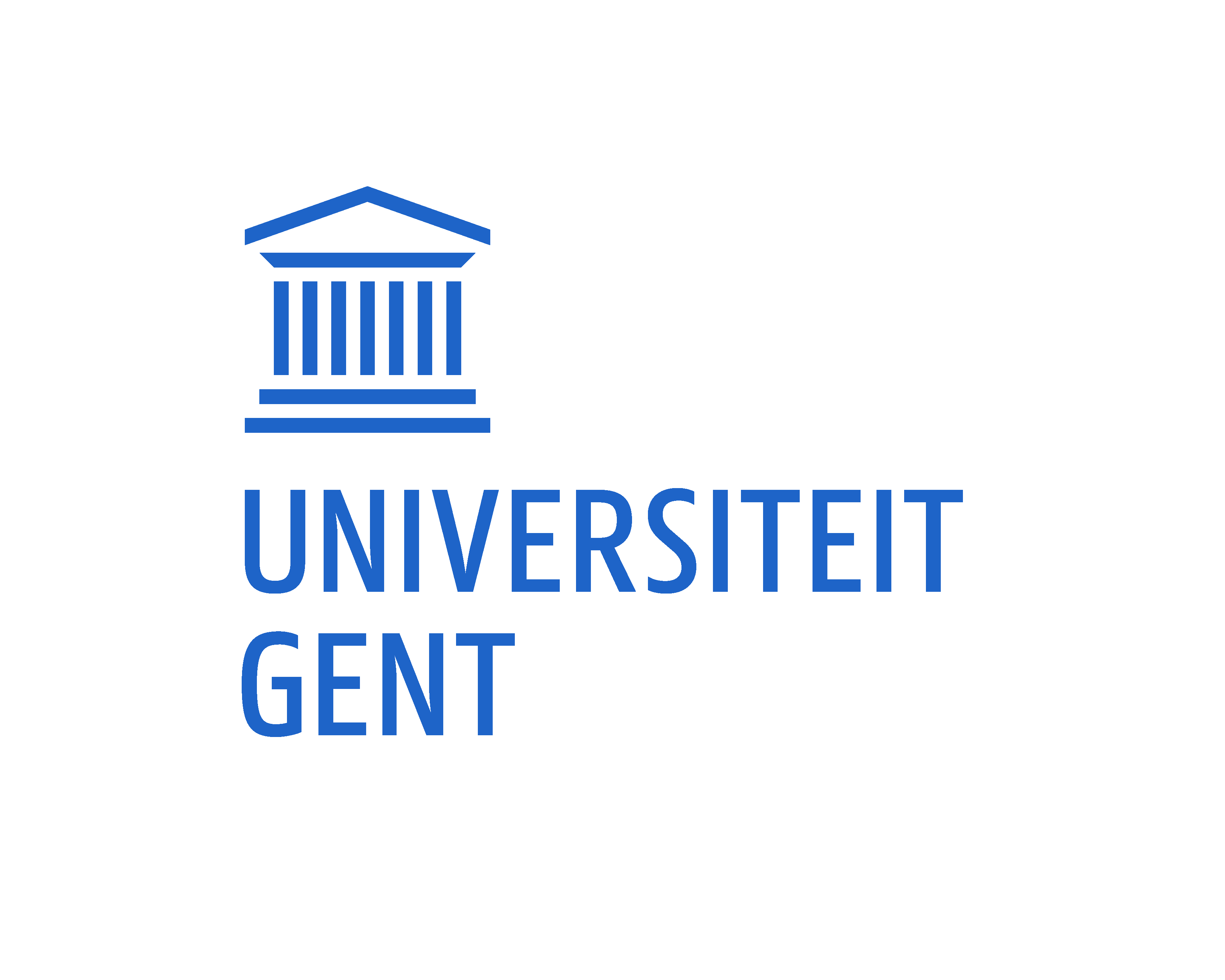}
\end{figure}

\begin{figure}[b!]
\includegraphics[scale = 0.1]{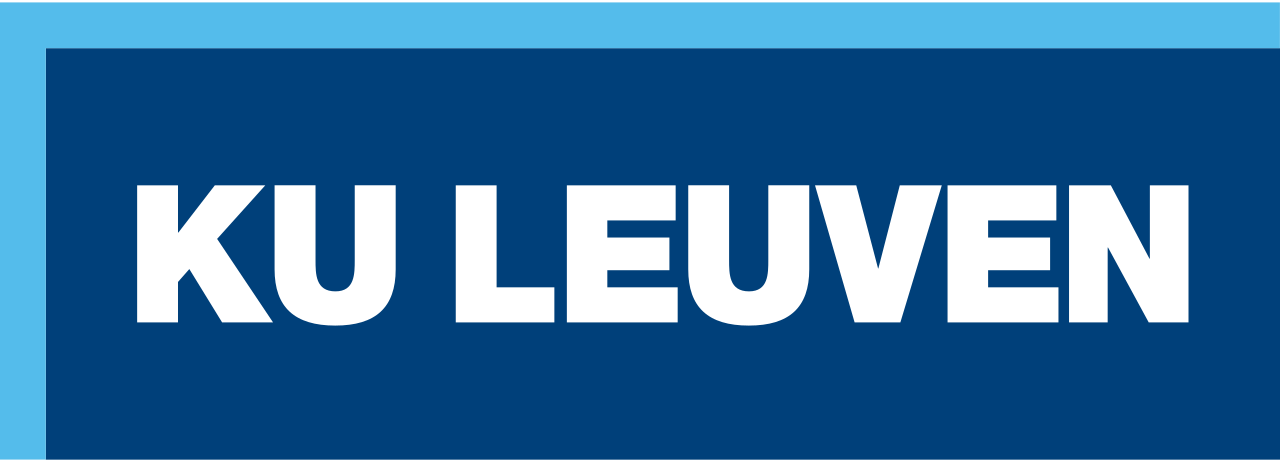}
\end{figure}

\end{flushleft}


\doublespacing

\section*{\uline{PERMISSION FOR USAGE}}
\addcontentsline{toc}{section}{\protect\numberline{}PERMISSION FOR USAGE}
``The author(s) gives (give) permission to make this bachelor's thesis available for consultation and to copy parts of this bachelor's thesis for personal use. In the case of any other use, the copyright terms have to be respected, in particular with regard to the obligation to state expressly the source when quoting results from this bachelor's thesis."

\vspace*{1cm}
\noindent{\color{ghent_blue} \textbf{STUDENT'S NAME: Fan Zhou}}

\vspace*{1cm}
\noindent{\color{ghent_blue} \textbf{STUDENT'S SIGNATURE:}}

\vspace*{1cm}
\noindent{\color{ghent_blue} \textbf{DATE:}} \hspace{2cm} 22-08-2022
\pagebreak

\section*{\uline{ABSTRACT}}
\addcontentsline{toc}{section}{\protect\numberline{}ABSTRACT}
The end result of the translation process is referred to as translationese (Gellerstam, 1986), also known as an interlanguage (Selinker, 1972) or the third language (Frawley, 1984). Its awkwardness, unnaturalness, and unidiomaticness set it apart from native-language texts in the target language since an extremely literal translation was utilized in an attempt to replicate the source texts' qualities. There are studies and research on reducing translationese to increase translation quality, and one of these is on translation relations, which encompasses a variety of translation techniques. Along with literal translation methods, there exist non-literal translation techniques that can manage translation that is incomprehensible by literal translation techniques.

\bigskip

\noindent This work identifies the distinctions between neural machine translation (NMT) and human translation (HT) based on translation relations. Seen as a type of metric for translation quality, the translation relation uses HT as a reference and assesses the translationese brought on by an NMT system, GoogleNMT (Wu et al., 2016). There are three research questions that need to be answered: how NMT and HT differ from one another in terms of the overall translation relations; how non-literal translation techniques are used differently by NMT and HT; and how NMT and HT differ from one another in terms of the factors that influence the use of specific non-literal translation techniques. 

\bigskip

\noindent Two parallel corpora are used: NMT-translated corpora as the experiment group and human-translated corpora as the reference one. Both parallel corpora have 9 genres with the shared source texts from the existing corpus (Zhai, 2020). Translation relations are manually annotated for aligned pairs with one side as the English token group and the other side as its corresponding translated Chinese token group. The use of translation relations in both parallel corpora is calculated for research comparisons, and in-depth linguistics-related knowledge which causes the use of translation techniques containing both semantics and syntax such as hyperonyms used by the translation technique of generalization and part-of-speech tagging alteration used by the technique of transposition is put to serve as a measurement for the application of translation relations. 

\bigskip

\noindent The final result shows that NMT uses more literal translation than HT by 17.41\% on an overall basis, and such overweight exists in each genre though there are differences in excess between genres. For the use of non-literal translation techniques, NMT shows the nearly- equivalent good performance with HT at the syntactic level especially when using translation techniques of lexical shift, transposition, and unaligned-reduction; however, its performance is not ideal when the translation using the techniques of particularization, figuration, equivalence, and generalization on the semantic levels.

\vspace{1cm}

\noindent\textbf{Keywords:}\quad Translation Relation, Translation Techniques, Translationese, Neural Machine Translation, Translation Quality, English-Chinese Translation
\pagebreak

\section*{\uline{ABBREVIATIONS}}
\addcontentsline{toc}{section}{\protect\numberline{}ABBREVIATIONS}
\textbf{acl} \hspace{2.2cm} clausal modifier of noun (adjectival clause)

\noindent \textbf{ADJ} \hspace{2cm} adjective

\noindent \textbf{ADP} \hspace{2cm} adposition

\noindent \textbf{ADV} \hspace{2cm} adverb

\noindent \textbf{ADV} \hspace{2cm} adverb

\noindent \textbf{advmod} \hspace{1.4cm} adverbial modifier

\noindent \textbf{amod} \hspace{1.8cm} adjectival modifier

\noindent \textbf{ASTrED} \hspace{1.5cm} aligned syntactic tree edit distance

\noindent \textbf{aux} \hspace{2.1cm} auxiliary

\noindent \textbf{AUX} \hspace{2cm} auxiliary

\noindent \textbf{BLEU} \hspace{1.8cm} bilingual evaluation understudy

\noindent \textbf{case} \hspace{2cm} case marking

\noindent \textbf{cc} \hspace{2.3cm} coordination

\noindent \textbf{CCONJ} \hspace{1.5cm} coordinating conjunction

\noindent \textbf{clf} \hspace{2.3cm} classifier modifier

\noindent \textbf{conj} \hspace{2cm} conjunct

\noindent \textbf{dep} \hspace{2cm} dependent

\noindent \textbf{DET} \hspace{2cm} determiner

\noindent \textbf{dobj} \hspace{2cm} direct object

\noindent \textbf{EBMT} \hspace{1.7cm} example-based machine translation

\noindent \textbf{GNMT/GoogleNMT} \hspace{0.5cm} Google neural machine translation

\noindent \textbf{HT} \hspace{2.3cm} human translation

\noindent \textbf{HTionese} \hspace{1.3cm} human translationese

\noindent \textbf{HUME} \hspace{1.9cm} Human UCCA-Based MT Assessment

\noindent \textbf{INTJ} \hspace{2cm} interjection

\noindent \textbf{LSTM} \hspace{2cm} long short-term memory

\noindent \textbf{mark} \hspace{2.1cm} marker

\noindent \textbf{MT} \hspace{2.4cm} machine translation

\noindent \textbf{MTionese} \hspace{1.3cm} machine translationese

\noindent \textbf{nmod} \hspace{1.9cm} noun compound modifier

\noindent \textbf{NMT} \hspace{2cm} neural machine translation

\noindent \textbf{nn} \hspace{2.3cm} noun compound modifier

\noindent \textbf{nsubj} \hspace{1.8cm} nominal subject

\noindent \textbf{NUM} \hspace{1.9cm} numeral

\noindent \textbf{PART} \hspace{1.9cm} particle

\noindent \textbf{PB-SMT} \hspace{1.4cm} phrase-based statistical machine translation

\noindent \textbf{pobj} \hspace{2cm} object of a preposition

\noindent \textbf{POS} \hspace{2cm} part-of-speech

\noindent \textbf{poss} \hspace{2cm} possession modifier

\noindent \textbf{prep} \hspace{2cm} prepositional modifier

\noindent \textbf{PRON} \hspace{1.8cm} pronoun

\noindent \textbf{PROPN} \hspace{1.6cm} proper noun

\noindent \textbf{punct} \hspace{1.9cm} punctuation

\noindent \textbf{QE} \hspace{2.3cm} quality estimation

\noindent \textbf{RBMT} \hspace{1.9cm} rule-based machine translation

\noindent \textbf{SACr} \hspace{2cm} syntactically aware cross

\noindent \textbf{SCATE} \hspace{1.8cm} Smart Computer-aided Translation Envrionment

\noindent \textbf{SCONJ} \hspace{1.8cm} subordinating conjunction

\noindent \textbf{SMT} \hspace{2.1cm} statistical machine translation

\noindent \textbf{SVMs} \hspace{2cm} support vector machines

\noindent \textbf{TER} \hspace{2.2cm} Translation Edit Rate

\noindent \textbf{WER} \hspace{2.2cm} word error rate

\noindent \textbf{X} \hspace{2.6cm} other

\noindent \textbf{YAWAT} \hspace{1.7cm} Yet Another Word Alignment Tool
\pagebreak


\newpage
\tableofcontents
\pagebreak
\pagenumbering{arabic}
\section{\uline{Introduction}}
Machine translation has been making significant progress in the past several decades, and one of the most important tasks is to achieve human parity which is defined by Hassan et al. (2018), which means the quality of machine translation reaches that of human translation, or even to reach the level of non-translated texts in target languages (Itamar, 1990; Toury, 2012; Baker, 1993, 1996, 2004; Dayrell, 2007). Although it is assumed unattainable at the current stage, there are various research and studies advancing the development in this direction. One research topic is about translationese reduction. In this dissertation, translationese in neural machine translation is detected with human translation as reference, measured by a metric named translation relations. 
\subsection{Choice of NMT system and its human parity}
There are three types of machine translation systems guiding the current machine translation domain, and they are rule-based machine translation, statistical machine translation, and neural machine translation. More details are introduced in the following part. \\
(1) Rule-based machine translation (RBMT) converts the source languages to target languages based on the linguistics-based rules specified by humans (Hutchins, 1995), featuring consistency and stability but low efficiency (Dove, 2012). Different languages have their own rules and contain oceans of text information; therefore, highly automation and efficiency are its two main problems. \\
(2) A corpus-based system with statistical machine translation (SMT) or phrase-based machine translation (PB- SMT) as the primary method uses machine learning algorithms by using a large amount of parallel corpus as input, overcoming the problem of using labor work to boost efficiency automatically (Lopez, 2008); however, it has issues such as with translation for idiomatic expressions, compound words that have to be translated by more than one word, long dependencies, and ambiguous words with different meanings depending on contexts (Nie{\ss}en, 2000). \\
(3) The currently predominant neural machine translation (NMT) system has gained the most extensive popularity in machine translation domain. In contrast to more established system SMT, NMT makes use of its architecture and capacity to capture complex sentence dependencies, which suggests that it has a great deal of potential to become a new trend in machine translation. Following the development of a simple model, several NMT models have been put out, some of which have made significant advancements and produced cutting-edge outcomes. (Yang et al., 2020). Though much improvement has been made, there are still problems of quality, especially rare words, long sentences, and word alignment (Koehn et al., 2017). \\
\\
Based on the previous rough analyses and comparisons for three machine translation systems, NMT has its advantages to be selected as the research target tool in this work, and its translation output is focused in this dissertation. Despite the advantages of the NMT system compared with other machine translation systems, how the translation quality by the NMT system is, and whether it achieve the human parity or the level of non-translated texts in target languages are questions discussed in this dissertation.  
\\
\\
A study assessing news translation from Chinese to English at the macro level claims that human parity has been reached (Hassan et al., 2018). However, the evaluation for translation quality at the document level shows a better preference for HT than for NMT (Laubli, 2018). Super-human translation performance by NMT between English and the other three languages, which are reassessed based on WMT 2019, is refuted (Toral, 2020). NMT shows worse performance when the sentences become shorter (Wan et al., 2022). 
\subsection{NMT translationese}
Against the overall backdrop of relative feeble translation quality produced by NMT systems, much work has been done on the integration of semantics and syntax into the NMT system, and there are significant improvements on the NMT quality. In this dissertation, Google Translate\footnote[1]{https://translate.google.com/} is used for NMT tool. Google Translate, also known as Google Neural Machine Translation (GoogleNMT or GNMT) is published by Google in 2016, which applies the example-based (EBMT) machine translation approach and uses artificial neural networks to increase translation fluency and accuracy. Instead of learning translations phrase by phrase, the GNMT network may do interlingual machine translation by storing the semantics of the sentence (Wu et al., 2016). Although large progress has been made by Google Translate or GNMT, there are still unreadable or unnatural translation produced by it. According to the study by Rosaria and Riana (2021), the translation is still wordy and incomprehensible, since literal translation is much used and text contexts are ignored. \\
\\
There are two examples to illustrate such problems:\\

\begin{center}
\setlength{\tabcolsep}{8.5mm}
\begin{tabular}{@{}ll@{}} 
\toprule

Source sentence & John can be relied on. He eats no fish and plays the game. \\
\addlinespace
Reference translation & \begin{CJK*}{UTF8}{gbsn}约翰为人可靠，既忠诚又正直。\end{CJK*} \\
 & (John can be relied on. He is loyal and trustworthy.) \\
\addlinespace
Google Translate & \begin{CJK*}{UTF8}{gbsn}约翰为人可靠，他不吃鱼，玩游戏。\end{CJK*} \\
 & (John can be relied on. He doesn’t eat fish but plays the game.) \\
\bottomrule
\end{tabular}
\end{center}

\begin{center} 
Table 1: Example-1 of machine translation
\end{center}

\begin{center}
\setlength{\tabcolsep}{7mm} 
\begin{tabular}{@{}ll@{}} 
\toprule
Source sentence & \begin{tabular}[t]{@{}l@{}}The track is aligned to industry-leading certifications, \\ which employers look for as validation of your exceptional skill.\end{tabular} \\ 
\addlinespace
Reference translation & \begin{tabular}[t]{@{}l@{}}\begin{CJK*}{UTF8}{gbsn}课程均与行业领先的认证直接挂钩，雇主会据此评判您是否具备卓越技能。\end{CJK*} \\ (Each course is aligned to industry-leading certifications, \\ according to which employers evaluate if you have exceptional skills.)\end{tabular} \\
\addlinespace
Google Translate & \begin{tabular}[t]{@{}l@{}}\begin{CJK*}{UTF8}{gbsn}发展道路均与行业领先的认证保持一致，雇主将其视为卓越技能的验证。\end{CJK*} \\ (Each track is aligned with industry-leading certifications, \\ employers consider it as validation of exceptional skills.)\end{tabular} \\
\bottomrule
\end{tabular}
\end{center}

\begin{center} 
Table 2 Example-2 of machine translation
\end{center}

\begin{CJK}{UTF8}{gbsn}In Example-1, “eats no fish and plays the game” is an idiom in English, should be paraphrased as “loyal and trustworthy” , and then translated into the corresponding target language Chinese as “既忠诚又正直” (loyal and trustworthy). However, Google Translate translates it literally without taking it as an idiom. While in Example-2: “track” is not necessarily translate into “道路” (road) but “课程” (course), and “look for as validation of” not translated but transposed and modulated into “据此评价是否” (according to which employers evaluate if). \end{CJK}

From the two examples above, the translation produced by Google Translate is of unnaturalness and unreadabilty. According to Gellerstam (1986), if the translated texts sometimes do not accord with the manner of the target language, however, with the “footprints” or trances of the source language, the linguistic phenomenon is called “translationese,” which causes translation problems such as being unnatural and hard to understand and violating habitual use of target languages. \\

Translation by Google Translate in the examples above can be seen as a kind of typical translationese of being replicating information from the source texts. Translationese is pervasive in the GNMT. In this dissertation, translationese is detected to assess the translation quality and performance by Google Translate.

\subsection{Translation relations}
Texts are prone to be literally translated through machine translation. In NMT, literal translation predominates, and Google Translate applies literal translation by 86.8\% (Sipayung, 2021). However, literal translation, or word-for-word translation (Newmark, 1981), may cause awkwardness or ungrammaticality of translation, and even translation errors for idioms and some other language expression patterns, which can be seen as translationese. \\
\\
In translation process by human translators, there are translation theories that guide the overall work of translation, and translation strategies and techniques are applied for specific translation process to ensure the translation quality. With these traditional guidelines, HT is thought to be more reliable than machine translations, and there is less translationese in HT than in MT. However, it is unclear that whether such a common sense is correct or not, and to what degree of these traditional translation guidelines are used in machine translations. Therefore, traditional translation guidelines are applied into this dissertation as one methodology for measuring translationese. For the convenience of calculation in this dissertation, only the translation relations, or translation techniques are used as one metric for translationese detection in GNMT. \\
\\
For translation relations which proposed by Chuquet et al. (1989; Zhai et al., 2018), it refers to different types of translation techniques distinguished from literal translation. Based on the theory of translation relations, Vinay et al. (1958; Zhai et al., 2018) first established the taxonomy of translation relations. There are generally four types of translation techniques: literal translation, equivalence (semantic overlapping ranges of reference in translation), non-literal translation, and unaligned translation (translation reduction or explicitation). In the category of non-literal translation, there are sub-types of translation techniques, including modulation which means maintain naturalness but change forms, transposition which involves a shift from one grammatical category to another with meaning unchanged, generalization which refers to broader and more general term in translations, etc (Chuquet et al., 1989; Zhai et al., 2018). As total, there are 14 categories in translation relations, and more details can be seen in the part of Methodology in Section 3. 

\subsection{Summary}
This dissertation analyzes translation obtained from an NMT system (GNMT) in comparison to human translations taking translation relation as the metric for translationese measurement. Specifically, this study examines how different HT and MT are from one another in terms of how translation relations are viewed; how much non-literal translation relations are used when translating using the NMT system; and what distinctions can be made between NMT and HT in terms of the variables that trigger the use of specific non-literal translation techniques. This dissertation is expected to serve as a basis for the further goal of optimizing NMT by reducing translationese from the perspective of translation relations to reach human parity or even the non-translated texts in target languages.\\
\\
To this end, source sentences in English obtained from the data set that was compiled by Zhai et al.(2018) are translated into Chinese through the GNMT system. In this way, there are two sets of parallel corpora of HT and MT with the source texts shared and target texts compared. In comparison, the aligned pairs which consists of token groups in English (the source language) and those their corresponding token groups in Chinese (the target language) are annotated with translation relations according to the translation relation annotation guidelines by Zhai et al. (2019). \\
\\
After this section of induction, the following section collects related literature and work covering translationese, its features, translationese in machine translation, methods to optimize transationese, as well as translation relations in Section 2. Methodology and data sets are described in Section 3, and detailed data analyses are presented in Section 4. Finally, Section 5 will conclude. \\

\pagebreak
\section{\uline{Related work}}
Some theories and previous work are applied into this work for theoretical basis, covering translationese's definition and features, metrics for measuring translation quality, and translation relations. 
\subsection{Translationese}
Translationese refers to elements that are inherent in translations and may even be inescapable, as opposed to badly translated versions that emerge from a lack of translation expertise (Tirkkonen-Condit, 2002). Translationese in NMT and HT is compared, and is one main topic in this dissertation. In this section, translationese is discussed concerning its definition, features, its occurrence in machine translation and the its existence in HT and MT.

\subsubsection{Translationese definition}

There is a wealth of evidence from translation studies indicating works that have been translated are ontologically distinct from those non-translated texts in target languages. Any language that has been translated into another might be thought of as having its own dialect, or "translationese". The concept of translationese or its equivalence are developed in the following studies.\\ 

Selinker (1972) proposed the “interlanguage” which is defined as the utterances different from those by native speakers of the target language when second language learners attempted to utter the sentences of the target languages from their native languages. Interlanguage can be viewed as a distinct language when second-language learners have a grasp of the meaning of their mother tongue or source language, as opposed to the particular target language. \\

Based on interlanguage theory, Frawley (1984) put forward the concept of the “third code”. He described the translation process as recodification from matrix code (source text) to target code (target text), and interlanguage is only a tiny part of the total information available in matrix code, called “semantic essence”. The conversion from the interlingual part from the matrix code to the target code needs the placement of semantic information of semiotic codes belonging to elements of the target matrix. Such semantic transfer is called the “third code", a totally new code derived from both matrix and target codes. The “third code” individuates itself with its logic. Frawley also argued that “fidelity” of the “third code” made translation unreadable, uninteresting, and unsuccessful.\\

The idea of a comparable corpus described by Baker (1995) as "two distinct collections of texts in the same language: one corpus consists of original texts in one language, while the other corpus consists of translations into that language from a particular source language or languages." The two corpora's domains, languages, periods, and lengths should be comparable. Based on the concept of comparable corpus, Gellerstam (1986) compared the Swedish texts written by native speakers and translations from English to Swedish, and found that the source language leaves certain “fingerprints” from the source texts on the target language in the translation process. This phenomenon is termed as “translationese”. In addition, Gellerstam suggests that it might be caused by a difference in genre, such as a higher prevalence of detective novels in the translated corpus than in the original corpus (detective stories being often translated from English into Swedish) (Gellerstam, 1986, 1996).\\

Toury (2012) focused on the effects of interference, the process by which a specific source language leaves distinctive marks or fingerprints on the translation in the target language, so that translations from different source languages into the same target language may be regarded as other dialects of translationese. 

\subsubsection{Translationese features}

Translationese is pervasive in both HT and MT, and features of translationese are looked into for better understanding. \\

(1) Overall features of translationese\\
Translationese is claimed to have some fixed linguistic features that tend to be shared in translation of different languages. Extralinguistic features in translational texts are excluded in the discussion of this dissertation. Translationese features are analyzed in several studies. Translationese features in translational English texts are identified with the reference of original English, and it shows that there are lower lexical density, fewer lemmas, and high repetition of high-frequency words (Laviosa, 1998). Baker (1993) proposed the universal features of translation or “translation universals” with more focus on the improvement of translations binding with more features on target language vis-{\`a}-vis source language. Specific translational features in translated texts have been contained in translation universal, including explicitness (Blum-Kulka, 1986), disambiguation and simplification (Vanderauwera, 1985), conventional ‘grammaticality' (Shlesinger, 1991), repetition avoidance (Shlesinger, 1991; Toury, 1991), etc., which might be examined with similar corpora (Baker 1996). Consistent disparities between translated and original English's lexical or syntactic structure without translators’ awareness were investigated. Olohan (2001) found that translations are more explicit than non-translations. The quantitative analysis experiment collects comments of intuitions on original writing texts, and concludes that lack of deviant or disturbing features was considered a mark of original writing and vice versa for translated language. Moreover, less unique items are another feature of translated texts (Tirkkonen-Condit, 2002). Santos (1995) sorted out several translationese features from the grammatical respect to syntax and tenses of Portuguese and English as research languages, and essential features are higher progressive, higher frequency of compact constructions, fewer tenses, and more locative details. Potential strategies have been used in translations such as explicitation and implications from Finish to English without ideological motivations (Puurtinen, 2003). \\

Apart from traditional analyses, the methods of using machine learning and deep learning models also reveal some traits of translationese. Some potential factors include n-grams, word form, lemma, Part-of-Speech (POS) tagging, and elements mixed as features input into support vector machines (SVMs) (Baroni, 2006). The study shows that the distinctive contrast between the translated and original texts is primarily affected by function words and POS n-grams. Ilisei et al.'s (2010) study suggests that the most relevant features that distinguish translations from untranslated texts are lexical richness, the proportion of grammatical words to lexical words, sentence length, word length, and some morphological attributes like nouns, pronouns, finite verbs, conjunctions, and prepositions by using a supervised learning technique. Two lexical features, namely high-frequency words and function words are used to identify translationese (Nisioi et al., 2013). For the supervised classification of translationese, it has been demonstrated that lexical and structural aspects of the text, such as characters, POS, and contextual function words, are useful. (Volansky et al., 2015). The unsupervised learning approach is used for translationese classification by using domain-based properties over translationese-related characteristics including function words, character trigrams, POS trigrams, contextual function words, and cohesive markers (Rabinovich et al., 2015). The study demonstrates the dominance of domain-specific properties over the characteristics of translationese. By observations and analyses for Hebrew, features at the word and sub-word level are selected as input features for translationese classification including token-based features (word unigram and function words), morphological aspects (Hebrew verbal patterns, status, progressive and prefix), n-grams, features that approximate word structure, and features above combined (Avner et al., 2016). \\

(2) Translationese features in translated Chinese texts\\
\begin{CJK}{UTF8}{gbsn}Characteristics of translationese in Chinese translated from English are summarized and analyzed. Function words, as translationese, are examined between statistical machine-translated Chinese and original Chinese. Generally, translationese in the translated Chinese texts lies in tenses, sentence kinds (questions or assertions), passive voice, genitive cases, and normalization. More specific phenomenon for translationese are discussed in some work. Kuo (2019) suggested that in machine-translated Chinese, there are overuses of “de" (的, “of” in English) which can be omitted in the adjective-noun combination, and noun phrases, “he”(和, “and” in English), “shi”(是, “is” in English), etc.; also there is under-use of non-existence type such as lái (来, connecting two verb phrases), adverb and preposition categories like yě (也,“also” in English), and postpositions such as děng (等, “etc.” in English). The use of third-person pronouns which are not commonly seen in non-translated Chinese texts was also found with higher frequency in the SMT texts (Kuo, 2019). Lin (2019) made a summary of Chinese translationese from the differences between English and Chinese on vocabulary including the range of meaning of a word, POS, and the word order. Language styles are also discussed in the study that original Chinese texts tend to use more verbs compared to the weakening of verbs in English, and barely use less logical relations with connectives; English commonly employs the passive voice, but Chinese virtually ever does, and Chinese sentences often utilize a positive voice and follow the standard sentence structure. Hu et al. (2018) use syntactic characteristics alone to successfully identify translated from original Chinese with the features of various forms of constituent and dependency parses; In Hu’s study, the translated Chinese, for instance, uses additional determiners, subject-position pronouns, NP modifiers including "的" and numerous NPs or VPs connected by the "、" which is a punctuation mark in Chinese. \end{CJK}

\subsubsection{Machine translation translationese}
With the rise of machine translation, its translation quality is a topic and many studies have done to improve it. Although there has been continuous optimization for machine translation quality, the translationese still exists in machine translation caused by different machine translation systems. Therefore, translationese caused by machine translation systems can also be called machine translationese. There are basically three types of machine translation systems, namely rule-based machine translation (RBMT), statistical machine translation (SMT), and neural machine translation (NMT). Different type has its own features in translationese. There are some work related to translationese in these three different machine translation systems. \\

(1) Rule-based machine translation
The traditional machine translation features rule-based machine translation (RBMT), also named the knowledge-based or linguistics-oriented approach (Hutchins, 1995), which converts the source languages to target languages based on the linguistics-based rules specified by humans. These rules are largely extracted from monolingual, bilingual, and multilingual dictionaries and grammars containing semantic, syntactic, and morphological regulations. 

For the translation quality, RBMT generally produces the target language featuring consistency and stability without relying on a large bilingual corpus as the translation process strictly abides by the stipulated regulations. RBMT outperforms both NMT and SMT for verb tense, aspect, and mood, as well as ambiguity, according to Burchardt et al. (2017)'s comparison of NMT to both SMT and RBMT in German-to-English and English-to-German. Most post-editing or translationese is caused by the second person forms and the deletion of redundant personal pronouns as subjects or possessives in the RBMT output. Fluency is a known problem for RBMT, especially in the case of Finnish, excessive usage of redundant pronouns, for instance, may cause the language to sound stiff and unnecessarily literal (Koponen et al., 2019).

(2) Statistical machine translation 
The fundamental concept behind statistical machine translation is to create a statistical translation model by statistically analyzing large amount of parallel corpora for training a model, and then to utilize this model for translation (Lopez, 2008). The phrase-based machine translation has replaced the earlier word-based translation, and has incorporates syntactic data to increase translation accuracy (Och et al., 2000). 

However, several studies show that there are translation problems for SMT system. Nie{\ss}en (2000) suggested that translation problems lies in idiomatic expressions, compound words that have to be translated by more than one word, long dependencies, and ambiguous words with different meanings depending on contexts. Compared to neural translations, phrase-based SMT has more translation errors in fluency and accuracy than NMT for English-Dutch parallel corpora according to Van Brussel et al. (2018). Besides, statistical engines provide clearer evidence of syntactic simplification produced by phrase-based SMT (Bizzoni et al., 2020).

(3) Neural machine translation
The state-of-the-art NMT approach is based on the transformer network (which is also an encoder-decoder approach). According to Wang et al.(2021), NMT has become the method of choice in the MT community as a result of these benefits. In the study of Aranberri (2020), the differences in translationese in the translation output produced by different MT systems (4 NMT systems and 1 RBMT system) are compared. Certain translationese features such as lexical variety, lexical density, length ratio, and POS sequence are analyzed, and the results show that NMT systems such as Google Translate and EJ-RBMT can render less translationese with these features as metrics compared with RBMT system. These features are called language-independent features; while Isabelle et al. (2017) made analyses of translationese on language-specific features, such as word concordance and insertion of words (De Clercq et al., 2021). In this study, machine translationese was examined by using the translation output produced by one SMT and three NMT systems. Five language-independent characteristics serve as the metrics for the differentiation between the original and machine-translated French, and the result shows that texts that have been machine translated frequently use the same word combination; the type-token ratios in machine translated texts are lower than reference non-translated texts (De Clercq et al., 2021). NMT performs best on a variety of features like coordination and ellipsis, multi-word expressions, long-distance dependencies, and named entities (Burchardt et al., 2017). The research by Toral et al. (2017) examined six languages from four different families (Germanic, Slavic, Romance, and Finno-Ugric), and they discovered that for all language directions outside English, the best NMT system outperformed the best phrase-based SMT system. Bizzoni et al. (2020) suggested that SMT produces over-simplified structures, while neural systems seem able to deal better with the complexity of their source. According to Jia et al. (2019), NMT delivers translations with greater quality than phrase-based SMT for both simple and complex texts.

Although their findings indicated that NMT systems have better performance than other machine translation systems, NMT performed badly when translating extremely long sentences (Toral et al., 2017). Recent error assessments reveal that NMT decreases many sorts of mistakes and generates more fluent output, particularly in morphologically rich languages like Czech, Finnish, Serbian, and Croatian (Klubi{\v{c}}ka et al., 2017; Toral et al, 2017). When comparing the number of necessary edits, NMT can be shown to minimize word order mistakes compared to both the SMT and RBMT outputs (Koponen et al., 2019). However, some research (e.g. Castilho et al. 2017) has suggested that fewer word order and word form errors increase the fluency of NMT, but not necessarily the adequacy and that this improving fluency may produce more misleading translations because NMT may produce grammatically correct sentences that do not match the meaning of the source text. However, according to Koponen et al. (2019), such reductions are lower in NMT(2.2.\%) than in RBMT (3.3\%) or SMT (2.7\%). In machine translationese feature analysis, transformer models perform better than phrase-based SMT (PB-SMT) and LSTM in terms of lexical and morphological diversity. The ratings for PB-SMT, LSTM, and transformer differ much less than those for original texts, demonstrating that the MT systems have a more detrimental effect on the morphologically richer languages (in terms of diversity and richness) (Vanmassenhove et al., 2021).

\subsubsection{Machine tranlationese v.s. human translationese}

Translationese exists in both MT and HT. Therefore, the comparison between MT and HT can also be seen as that between machine translationese and human translationese. There are studies showing such comparisons. Popel et al. (2020) suggested that CUBBITT model which is deep-learning based fared noticeably better than professional agencies translating English to Czech news in terms of maintaining text meaning (translation adequacy) when translation adequacy, fluency, and overall translation quality of both MT and HT are inspected by evaluators. However, some other studies show that MT is overall worse than HT in terms of translationese comparison . Automatic classification between MT and HT based on lexical diversity was done by Fu et al. (2021) with the factor of accuracy above the chance level. Translationese across human and machine translations from text and speech is compared by using POS perplexity and dependency length to conclude that machine translation can over-correct translationese effects, not following the characteristics of training data. Also, the neural systems' translations were found often more sophisticated yet near to the HTs, or to the originals, and machine translation usually exhibits higher degrees of structural interference from source languages and lower levels of adherence or over-adherence to the target language’s standards than human translation (Bizzoni et al., 2020). Vanmassenhove et al. (2019) found that compared to the text produced by humans, the MT process often results in a loss in lexical diversity and richness. Several observations on MT output compared to HT from English to Dutch are that a significant number of the MT translated sentences were flawed; there were less lexical variety and local coherence in NMT; NMT are more likely to mimic the source sentence's grammatical structure; besides, the style of HT differs from that of MT (Webster et al., 2020). 
\subsection{Metrics for translation quality}

Translation metric is a kind of method to evaluate the translation quality. In the past decades, many metrics have been developed for translation evaluation. There are many translation metrics nowadays containing automated metrics and human metrics. Although Chatzikoumi (2020) claimed that two types of metrics are in practice not distinguishable as human labor work and machine calculation are interwoven, the pros and cons of each metric are analyzed in the following section.

\subsubsection{Automatic evaluation of machine translation}
Automated metrics for evaluating translation quality are convenient, relying on computational calculation to reduce labor work and improve efficiency. Reference-based translation metrics is one type which includes some measuring approaches based on the ideas of edit distance and precision and recall (Chatzikoumi, 2020). There is Levenshtein’s distance which focuses on the differences of sentences on the character level by calculating the sum of edit operations from one sentence to another (Levenshtein, 1966), and it is developed into concentrating on words rather than character (EuroMatrix, 2007). Based on the idea of Levenshtein’s distance, there are Word Error Rate (WER) (Nie{\ss}en et al., 2000) and Translation Edit Rate (TER) (Snover et al., 2006) to calculate differences between translation and its reference sentences. The most often employed precision-based metric is the bilingual evaluation understudy (BLEU) (Papineni et al., 2002) which compares adequacy and fluency between reference sentences and translated output by using word n-gram with the score of from 0 to 1 as a measurement of closeness. However, there is no differentiation between translations with extremely low scores in low quality or highly free translations (Coughlin, 2003), and the failure to distinguish subtleties (Lavie, 2010). The use of lemmatization, synonymization, stemming, and paraphrasing is another kind of method that is different from the reference-based approach (Callison-Burch et al., 2006). Different from reference-based metrics, quality estimation (QE) metrics can classify translation into a good or bad translation without references. It mainly extracts features from source and target sentences from the perspective of complexity, fluency and adequacy, concerning the calculation of tokens, POS tagging, dependency relations, etc. (Specia et al., 2013). Besides, there are some other methods and metrics optimized and developed. CHRF (Popovi{\'c}, 2015) utilizes character-level n-gram to identify morphological differences. Lo (2019) determined semantic similarities of phrases in translation output with reference, namely YISI-1, and there is also its variant YISI-2. There are some other metrics involving more linguistic information. Popovic et al. (2011) calculate the similarity scores of source and target (translated) sentences based on the probabilities of morphemes, 4-gram POS, and lexicon. The bilingual phrase tables' synonyms and the POS data from the matching task were used to create the TESLA assessment measures by Dahlmeier et al. (2011). Vanroy et al. (2021) combined dependency relations with word reordering to calculate word cross nodes, called SACr. They also proposed aligned syntactic tree edit distance (ASTrED) which aligns source and target dependency tree and calculate the edit distance between dependency parsed trees. Named entity knowledge is drawn from the literature on named-entity recognition, which seeks to recognize and categorize atomic elements in a text into distinct entity categories, to capture the semantic equivalence of sentences or text fragments (Marsh et al., 1998; Guo et al., 2009). Neural networks is used on translation quality assessment for pair-wise modeling to compare potential translations with a reference and select the best hypothetical translation by combining syntactic and semantic information into Neural networks (Guzm{\'a}n et al., 2017). The BERTScore metric was proposed by Zhang et al. (2019) and is based on contextual embedding. It gets around the typical drawbacks of n-gram-based metrics (such as synonyms and paraphrases), allows translations that are different from the references, and can evaluate the accuracy of a translation model in a contextual embedding space.

\subsubsection{Traditional metrics}
There are no absolute standards of translation excellence; there are only translations that are more or less appropriate for the context in which they are employed, according to Sager (1989). The establishment of such external criteria of translation quality is challenging, thus it is typical to choose a more constrained approach, focusing on the intrinsic features of translated texts and on translation errors as a means of measuring quality (Secar{\u{a}}, 2005). Many ideas have been proposed to characterize translation assessment in the past century. According to Nida (1964), the translation should be “dynamic” to conform to the different target audiences, and good translations are those in which the reader's response coincides with that of native speakers of the source language. Vermeer (2021) proposed the “skopos theory”, and he stressed that each text is created with a certain purpose, and when the purpose is fulfilled, the translation is a good translation. Peter Newmark (1981) distinguished between communicative translation (word-for-word) and semantic translation (sense-for-sense). According to him, communicative translation is biased toward the target language, free, and idiomatic, whereas semantic translation is biased toward the source language, literal, and loyal to the original material. The aim of a semantic translation is to preserve the exact contextual meaning of the original by adhering as closely as possible to the semantic and syntactic structures of the source language. The objective of a communicative translation is to have the same impact on the audience that the original had on its readers (Newmark, 1988). Lauscher (2000) proposed equivalence-based and function-based approaches for translation quality assessment. \\

However, the theories above are elusive for translators, and specific and standard metrics are explored. 22 types of translation errors are stipulated by American Translators Association ranging from terminology and register to accents and diacritical marks (Secar{\u{a}}, 2005). Point scale schema is designed to measure accuracy and fluency, and there are five-point scales (Callison-Burch et al., 2007) and seven-point ones (Przybocki et al., 2009). However, due to anomalies in the five-point scale, the adequacy and fluency metrics have been completely abandoned in the Workshop on Statistical Machine Translation assessments (Bojar et al., 2016). Human UCCA-Based MT Assessment (HUME), a semantic evaluation measure suggested by Birch et al. (2016), is another method of evaluating adequacy. Appraise (Federmann, 2010) is an open-source application that allows users to conduct MT assessment annotation activities concerning ranking. Direct Assessment involves the expression of a judgment of the quality of the MT output in a continuous rating scale (Graham et al., 2015), and it can capture the extent to which one translation is better than another rather than ranking interval-level scales (Graham et al., 2013). It is often used along with post-editing (Bentivogli et al., 2018). Tezcan et al. (2017) proposed a translation error identification guideline which specifies accuracy errors such as addition, omission and the untranslated and fluency errors like word form and word order in grammatical level and spelling in orthographic level. Based on it, an annotation tool named Smart Computer-aided Translation Environment (SCATE) where accuracy errors such as and fluency errors such as can be annotated is proposed.

\subsection{Translation relations}
No matter in MT or HT, there is always translationese in the translated texts which has discrepancies to original non-translated texts in target languages or human parity. To reach Human parity or even non-translated texts in target languages is the ultimate phrase in the translation process after the translationese reduction. The application of translation relation is one of good approaches to narrow such discrepancies and help reduce translationese in HT and may in MT. Before translation relation is applied to reduce translationese in MT, it can be checked and detected in MT to see how and how much it is used in MT.

\subsubsection{Translation relations}
According to Krings (1986), translation strategies were defined as a “potentially conscious plan for solving translation problems”. He built a tentative model of the translation process covering translation problems and translation strategies as solutions. For example, if the problem is not of “potential equivalence”, “retrieval strategy” would be used. Van Leuven-Zwart (1990) investigated translational norms, methods, and strategies adopted by translators, and concluded that “generalization” is less frequently used than “specification”, and source language oriented translation is more than target language oriented translation. The strategy of foreignization was explored and found in dominance in the translation of cultural texts (Kwieci{\~n}ski, 1998). Translation strategies "involve the fundamental tasks of picking the foreign material to be translated and establishing a technique to translate it," according to Venuti (2000). He uses the terms "domesticating" and "foreignizing" to discuss translation techniques. Strategies are "a collection of skills, a set of activities or processes that favor the capture, storage, and/or exploitation of the information," according to J{\"a}{\"a}skel{\"a}inen (1999). According to him, strategies are "flexible and heuristic in character and their adoption indicates a decision affected by revisions in the translator's aims.” Newmark (1998) summed several translation methods or strategies from the whole text or the macro level including word-for-word translation, literal translation, faithful translation, semantic translation, adaption, free translation, idiomatic translation, and communicative translation. \\

When translation strategies and methods are applied to the translation process, translation tactics or techniques are worth mentioning. There are five features of translation techniques: “affecting translation output, classified by comparison with the original, affecting micro-level unit, discursive and contextual, and functional” (Molina et al., 2002). As Newmark (1998) mentioned that translation process or “procedure” is within “the sentence or the smaller units of language”. He classified different translation procedures covering transference, neutralization, cultural equivalent, functional equivalent, descriptive equivalent, componential analysis, synonymy, through-translation, shifts or transpositions, modulation, recognized translation, paraphrase, couplets, and notes. Molina et al. (2002) specified literal translation based on Newmark’s translation procedure, and they defined it into three sub-categories including borrowing, claque, and literal translation; transposition, modulation, equivalence, and adaption belong to the category of oblique translation which happens when the word-for-word translation is impossible. They also summed seven opposing pairs as Explicitation vs. Implicitation, Generalization vs. Particularization, and Reinforcement vs. Condensation. Based on all these translation techniques, translation relation, a kind of interlingual relation, was proposed to categorize translation techniques aside from literal translation (Vinay et al,, 1958; Chuquet et al., 1989; Zhai et al., 2018). Vinay et al. (1958; Zhai et al., 2018) were the ones who originally postulated and created a taxonomy of translation procedures. Based on theories that Chuquet et al.’s (1989; Zhai et al., 2018) work has clarified, there is a hierarchy of translation relations. This hierarchy of translation relations is used in this dissertation as a guideline, and more details are to be declared in Section 3. 

\subsubsection{Non-literal translation in NMT}
There are researches and studies into translation techniques that are displayed in the machine-translation output. Machine translation systems are prone to use more literal translation technique which may cause stiff, unnatural and even wrong translation. Machine translation with “literal” translation as the dominance in the 1990s for idioms was proposed as inappropriate (Volk, 1998). Literal translation by machine translation systems was thought to be notorious when they translated idioms (Hutchins, 1995). Literal translation mistakes are thought to represent a significant category of idiom-related translation problems (Manojlovi{\'c} et al., 2017; Shao et al., 2018). \\

Therefore, non-literal translation techniques applied in machine translation or specifically in NMT are more focused in this dissertation. Non-literal translation exists because it is standard procedure in translation for the translator to select a non-literal translation in light of factors like naturalness, discourse coherence, and other factors (Deng et al., 2017). Non-literal translations produced by humans reflect the diversity of human languages and are occasionally necessary to maintain accuracy and fluency (Zhai, 2020). For translation relations used in NMT, most studies are on idiom translation problems caused by literal translation. The fact is that most of idioms are translated with non-literal translation techniques, but according to the evaluation for the language model of Transformer of NMT (Dankers et al., 2022), Transformer’s tendency to process idioms as compositional expressions contributes to literal translations of idioms. In order to make the idiom translated correctly, Fadaee et al. (2018) retrieved sentences containing idioms and mixed them with non-idiom phrases as a training data set to train the NMT. Apart from idioms, some other literal translation problem are studied. For example, Gamallo et al. (2021) suggested that the large amount used of literal translation causes transferring the constructions of the source language to the target language, and they proposed to use monolingual corpora instead of parallel ones with unsupervised translation can make the hybrid SMT plus NMT system produce more non-literal translation output on passive voice structured sentences. \\

According to studies mentioned above, most studies only concentrate themselves on one translation problem (idiom translation) caused by using large amount of literal translation. However, there is few work involving the all translation techniques in NMT. Apart from literal translation, there are several non-literal translation techniques that can be analyzed and detected in NMT to see how these non-literal techniques are used in NMT. In this case, this dissertation focuses on the application of both literal translation and specific non-literal translation techniques in NMT.

\pagebreak
\section{\uline{Methodology}}
In this section, some methodologies\footnote[2]{Data and codes available at https://github.com/louisefz/translation\_relations .} which applied into this dissertation are introduced, including translationese comparison, translation relations as an annotation guideline, and some language-independent and language-specific metrics for supplementary measurement. 
\subsection{Translationese comparison}
This part introduces the used data, how the data is formed into machine translationese (MTionese) and human translationese (HTionese), and the comparison between MTionese and HTionese. 

\subsubsection{Data source}
The experiment is based on the comparison between MT and HT, and therefore, there are two parallel corpora which consists of a set of texts in the source language and their translations in the target language: one is the source text and its translation by human translators which is defined as HT data, and the other is the source text which is shared with that of HT data and its translation created by Google Translate\footnote[3]{The translation in this study is produce in May, 2022 by Google Translate (https://translate.google.com/).}, which can be called MT data. For HT data, both source and target data are from Zhai et al. (2020). This existing data is from a new corpus of bilingual parallel texts they built, which covers eleven genres of texts including art, literature, law, material for education, microblog, news, official document, spoken, subtitles, science, and scientific article. The translation direction is all from English to Chinese for each genre except for the genre of scientific articles whose translation direction is from Chinese to English. However, for this kind of genre, I have changed the direction of the translation so as to ensure the target language is all Chinese, making it consistent for both parallel corpora. The source of the existing data from Zhai et al. is sampled from several different bilingual corpora which have their own features are employed covering UM-corpus (Tian et al., 2014), UT-corpus (Liu et al., 2015), UB-corpus (Chang, 2004), UnitedNations-corpus (Ziemski et al., 2016), and online bilingual journals (Zhai et al., 2020). For genres of news, literature, art, scientific article, and official documents, they are from the corpora of UT-corpus, UB-corpus, UB-corpus, own construction, and UnitedNations-corpus; as to the law, material for education, microblog, spoken, subtitles and science, they are collected from UM-corpus. The newly-built corpus has a total of 2200 pairs of bilingual sentences with high-quality human translation. 

The work in this dissertation uses the data from the experiment of automatic classification for translation relations from English to Chinese by Zhai et al. (2020), and there are 9 genres covering education, law, scientific article, microblog, news, official document, science, spoken and subtitles. Each genre contains 50 sentences except for that of microblog which has 53 sentences, and there are 453 pairs of sentences in total. For MT data, the source texts in English are the same as the source data of HT data. 

After HT and MT data is prepared, both its source sentences and translation are tokenized. As for HT data, both source and target sentences has been tokenized by Zhai et al. (2020). Standford Tokenizer is used for tokenization of source texts in English, and THULAC (Li et al., 2009) is for tokenization of translation in Chinese. Since the tokenizer is not 100\% accurate, there are manual corrections by the annotator when checking and reviewing data. However, there are still some minute mistakes as the Chinese language, different from English, is more complicated for tokenization. Therefore, on the prerequisite of abiding by the tokenization principle of the Chinese language, there is some degree of intuition by the annotator.

\begin{center}
\setlength{\tabcolsep}{7mm}
\begin{tabular}{@{}lccc@{}} 
\toprule
\makecell[l]{Genre} & \makecell{Source token number} & \makecell{MT token number} & \makecell{HT token number} \\
\midrule
education & 1021 & 981 & 999 \\
laws & 1857 & 1681 & 1648 \\
microblog & 1108 & 1160 & 1139 \\
news & 1573 & 1424 & 1398 \\
officialDoc & 1713 & 1538 & 1485 \\
science & 975 & 914 & 880 \\
scientificArticle & 1367 & 1300 & 1288 \\
spoken & 703 & 666 & 691 \\
subtitles & 599 & 592 & 526 \\
Total & 10916 & 10256 & 10054 \\
\bottomrule
\end{tabular}
\end{center}
\begin{center} Table 3 Token number of each genre in source texts, machine translated texts, and human translated texts
\end{center} 

The source text in English has 10916 tokens, while HT Chinese and MT Chinese texts have 10054 and 10256 tokens respectively. More details of token number of each genre shows in Table 3. Overall, the number of English tokens is more than that of the rest two Chinese data sets, which is a regular phenomenon. However, the HT Chinese tokens are less than MT Chinese tokens except for the genres of education and spoken.

\begin{figure}[H]
\centering
\includegraphics[scale = 0.47]{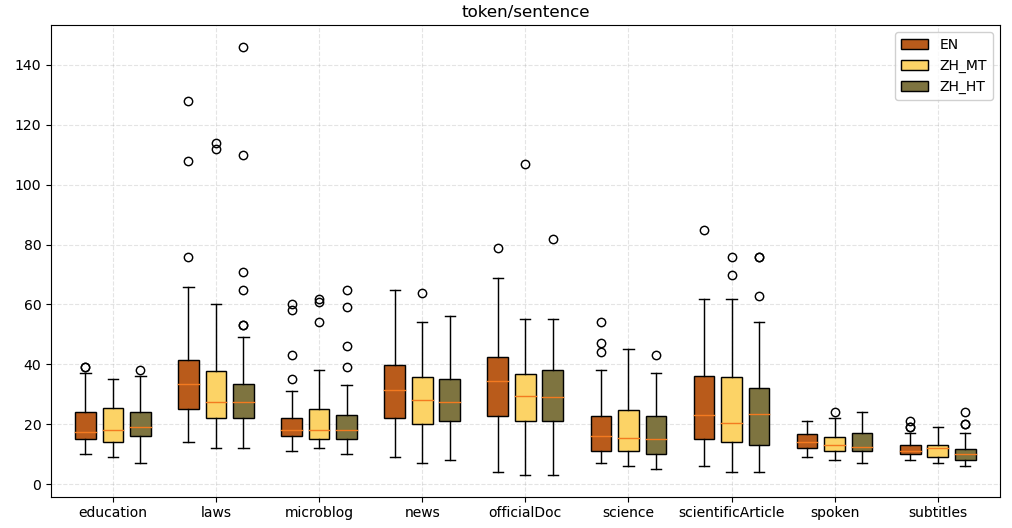}
\caption{Figure 1: The number of tokens in each sentence of each genre in three corpora}
\end{figure}

The number of tokens in each sentence is calculated, and Figure 1 reflects the distribution of tokens in each sentence of different genres and data sets. Three data sets in all are evenly distributed in each genre. For genres of laws, news, officialDoc, and scientificArticle, the sentence lengths are not stable, and these genres contain more long sentences. 

\subsubsection{MTionese v.s. HTionese}

There are two sets of data, one is HT data, and the other is MT data. The translation in both HT data and MT data can be seen as translationese as both of them have discrepancies from the non-translated texts in the target language. Therefore, there are machine translationese (MTionese) and human translationese (HTionese). In this work, these two types of translationese are compared to find more if NMT reaches the human parity in terms of translation relations. 

Translation relations which referes to the relations between different translation techniques can be equivalent to translation techniques. The application of translation relations into translation process can optimize the translation quality and reduce translationese. Concepts on translation relations and translation techniques are clarified in the following part. In this dissertation, the translation relation or technique is seen as a metric to measure translationese in both MT and HT data. The use of translation relations in the translation of HT and MT data is compared to see whether the performance of using translation relations in MT data is better, worse than, or equivalent to that in HT data. 

\subsection{Translation relations}
Translation relations or translation techniques can help improve translation quality and reduce translationese in the process of translation. In this part, translation relations as a metric for measuring translationese in both HT and MT data and as a guideline of annotation will be specified. 

\subsubsection{Details of translation relations}

Translation relation, a kind of interlingual relation, is defined to categorize translation techniques aside from literal translation (Vinay et al., 1958; Chuquet et al., 1989; Zhai et al., 2018). A taxonomy of translation processes was first proposed and built by Vinay et al. (1958; Zhai et al., 2018). The hierarchy of translation relations that is provided here is based on hypotheses that Chuquet et al.’s (1989; Zhai et al., 2018) work has elucidated. 

There is a hierarchical architecture of translation relations (see Figure 2) which has four types of translation techniques, namely literality, equivalence, non-literality, and unalignment. Beside, non-literal techniques have sub-categories of techniques, and the same for unalignment node. 

Nodes that are blue colored are translation techniques to be looked into in this dissertation. As for other white nodes, they are either more general to conclude some translation techniques or more specific to explain how their higher nodes work. The definition for each translation technique used in the dissertation will be given (Zhai et al., 2019) and exemplified.

\begin{figure}[H]
\centering
\includegraphics[scale = 0.47]{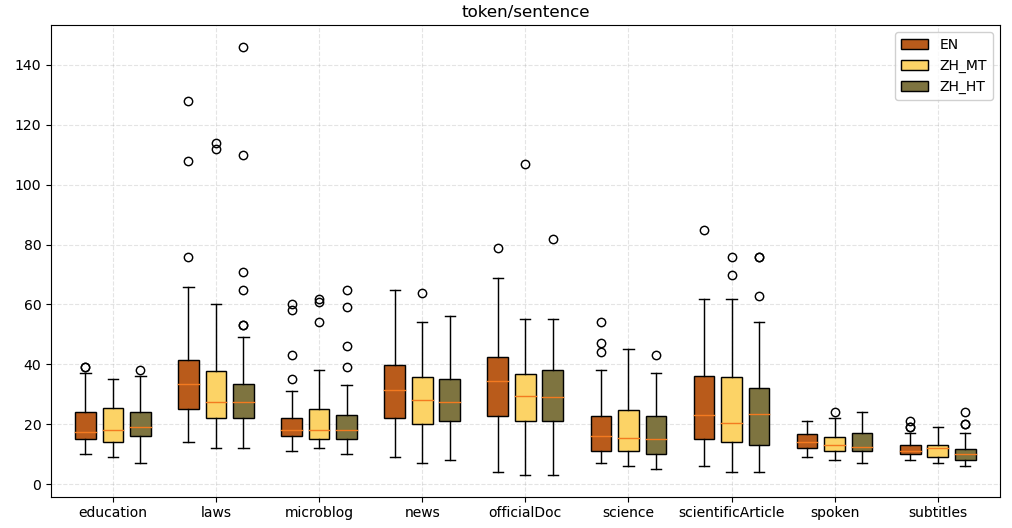}
\caption{Figure 2: Hierarchy of translation relations (Chuquet et al., 1989; Zhai et al., 2018)}
\end{figure}

(1) Literal translation is a type of direct translation or word-for-word translation that occasionally includes loan words or borrowing words from the source languages, the direct translation of idioms, the use of corresponding expressions when an absolutely literal translation is incomprehensible, the addition or deletion of determiners, the switching of singular and plural forms, etc.\\
Examples of literal translation:\\
literal a bronze ring \begin{CJK}{UTF8}{gbsn}→ 一 个 青铜 戒指\end{CJK}\\
As a husband, he is affectionate. \begin{CJK}{UTF8}{gbsn}→ 作为 一个 丈夫 ，他 十分 地 深情 。\end{CJK}

(2) Equivalence is the non-literal translation of proverbs, idioms, or fixed phrases; it is also semantic equivalence at the supra-lexical level, translation of words, and there is no change in meaning or point of view, even though the translator has generated a different translation.\\
Examples of equivalence:\\
There is no use in crying over the split milk \begin{CJK}{UTF8}{gbsn}→木已成舟\end{CJK} (The boat has been done. )\\
protect all locations \textbf{at all times} → \textbf{\begin{CJK}{UTF8}{gbsn}日夜\end{CJK}} ("day and night") \begin{CJK}{UTF8}{gbsn}保护 所有 的 地点\end{CJK}

(3) Modulation is a technique used by translators to keep the message as natural-sounding as possible by employing different forms and switching points of view. When literal translation would provide difficult or unnatural translation, this method is typically employed by translators. It involves shifting the viewpoint in order to get around a translation problem or to show how the target language speakers see the world. The semantic divergence between the source text and the target text may be caused by this category. All other phenomena, except its two sub-types of Particularization and Generalization, are represented by the sub-type. \\
Examples of modulation:\\
\begin{CJK}{UTF8}{gbsn}It is \textbf{difficult} . → 这 \textbf{不 简单 (not easy)}。\\
I like the dreams of the future \textbf{better} than the history of the past . → 我 \textbf{不 (not)}缅怀 过去 的 历史 ，\textbf{而 (instead)} 致力于 未来 的 梦想 。\\
\textbf{Maybe} we should postpone till the weekend . → \textbf{不如 (why not)} 我们 推迟 到 周末 吧 。\end{CJK}

(4) Particularization is to put the expression of translation in a clearer or more specific way. The translator would pick one of several target segments with a more precise meaning among those that may be created from the source segment. It occasionally explains a segment's context-specific significance. A pronoun in one sentence can be translated into the items that it refers to. \\
Examples of particularization:\\
\begin{CJK}{UTF8}{gbsn}" Yes , \textbf{put} you to bed , " she added lightly → " 是的 ， \textbf{服侍 (serve)}你 上床 睡觉 , " 她 小声 补充 说\\
He then requested \textbf{her} to stay where she was → 他 先 让 \textbf{苔丝 (Tess)} 在 外面 等 着\end{CJK}

(5)Generalization is the "conscious or unconscious semantic loss of one or more distinct senses" in translation (notional or pragmatic). The translator employed the more broad target term or phrase, which might be translated from a variety of source words or expressions. Sometimes non-fixed terms are used to refer to idioms, and the figurative picture is taken away from the source languages in the translation. \\
Examples of generalization.\\
\begin{CJK}{UTF8}{gbsn}a research that will be embraced by \textbf{millions of} bleary-eyed Britons → 一 项 即将 被 \textbf{广大 (many)} 睡眼惺忪 的 英国人 所 知道 的 研究\\
But should \textbf{tempers} rise in the Middle East, the price could jump again. → 如果 中东 \textbf{战火 (wars)} 重 燃 的 话，无疑，油价 会 再度 上涨 。\end{CJK}

(6) Transposition is the alteration of the order of grammatical categories or parts of speech without altering the meaning. \\
Examples of transposition:\\
\begin{CJK}{UTF8}{gbsn}Those who are \textbf{experienced} have feeble imagination . → 有 \textbf{经验(experience, noun)}的 人 缺乏 (lack, verb) 想象力 。\\
people \textbf{of} Iran → 伊朗 \textbf{的 (’s, particle)}人们\end{CJK}

(7) Modulation plus Transposition contains any sub-categories of modulation plus transposition combined. \\
\begin{CJK}{UTF8}{gbsn}Examples of modulation plus transposition:\\
Tobacco companies are barred from running cigarette ads in popular teenager magazines and \textbf{from targeting youth} . → 烟草 公司 不准 在 青少年 普遍 阅读 的 刊物 上 刊登 香烟 广告 ，不准 \textbf{以 青少年 为 拉拢 对象} 。\end{CJK}

(8) Figurative translation is to use an idiom or a metaphor to translate a non-metaphor or a non-fixed term, and it sometimes uses personification. \\
Examples of figurative translation:\\
\begin{CJK}{UTF8}{gbsn}pressured the people a little bit about it → 刨根问底 ("inquire into the root of the matter”)\\
For Joanne, new opportunities are \textbf{opening} . → 对 乔安娜 而 言 ，新 的 机遇 现 已 向 她 \textbf{招手 (opportunities are waving hands to her)}。\end{CJK}

(9) Lexical shift is that the message remains unchanged despite the translation's literal inaccuracy. They are only slight vocabulary modifications that don't require any translation techniques.\\
Examples of lexical shift:\\
\begin{CJK}{UTF8}{gbsn}include the following additional \textbf{responsibilities} → 包括 下列 新增 的 \textbf{职责 (plural noun to the singular)}\\
He also \textbf{indicated} that the United States will hold negotiations with Cuba → 他 还 \textbf{表示 (past tense to present tense)}， 美国 将 与 古巴 举行 谈判\end{CJK}

(10) Translation error is errors that occur in translation. \\
Examples of translation error:\\
\begin{CJK}{UTF8}{gbsn}database \textbf{connection} method → 数据库 \textbf{访问 (access, should be “连接”)}方式\end{CJK}

(11) Uncertainty is one category that no translation categories mentioned above can be assigned to the translation. 

(12) Other types: Apart from the eleven categories which can be alignment on both source and target languages sides, there are tokens that are not aligned on either side. If there are tokens unaligned from the source side, this case is ascribed to the category of “reduction”; the tokens which are not aligned on the target side are labeled with “explicitation”. There is also another type named “no type” which means the untranslatability of functions words and segments that were not translated but had no bearing on the message. \\
Example of reduction:\\
\begin{CJK}{UTF8}{gbsn}Peter \textbf{is} six years old. → 彼得 六岁。\end{CJK}\\
Example of explicitation:\\
\begin{CJK}{UTF8}{gbsn}the knife → 这 \textbf{把 (Chinese measure word)}刀\end{CJK}\\
Example of no type:\\
\begin{CJK}{UTF8}{gbsn}The tragedy of the world is \textbf{that} those who are imaginative have but slight experience. → 世界 的 悲剧 就 在于 有 想象力 的 人 缺乏 经验。\end{CJK}

Definitions for each translation techniques and examples above are either from Zhai’ (2019) annotation guideline of Translation Techniques for English-Chinese or corpora data.

\subsubsection{Annotations of translation relations}
When all sentences are tokenized, tokens of the source and target sentences are aligned, and Tsinghua Aligner is used for automatic token alignment so as to reduce the labor alignment. According to the user manual of Tsinghua Aligner (Liu, 2015), once the source and target texts are entered, there will be an alignment file with the token index aligned. For example, in Figure 3 there are two sentences in both source.txt (Chinese pinyin) and target.txt respectively. Line 2 is the source sentence, corresponding to line 6, the target sentence, and the same with line 3 to line 7. In Figure 4, Line 1 is the token index correspondence between line 2 and line 6 in Figure 3, and line 2 is for line 3 and line 7 in Figure 3; in “1-0”, “1” represents “he” (both) in line 2 of Figure 3, and “0” represents for “both” in line 6 of Figure 3. However, the automatic word alignment tool cannot ensure the alignment outcomes are all correct. Therefore, checking for alignment several times is needed when the annotator annotates the translation relations on the platform named YAWAT (Yet Another Word Alignment Tool) (Germann, 2008).

\begin{figure}[H]
\centering
\includegraphics[scale = 0.47]{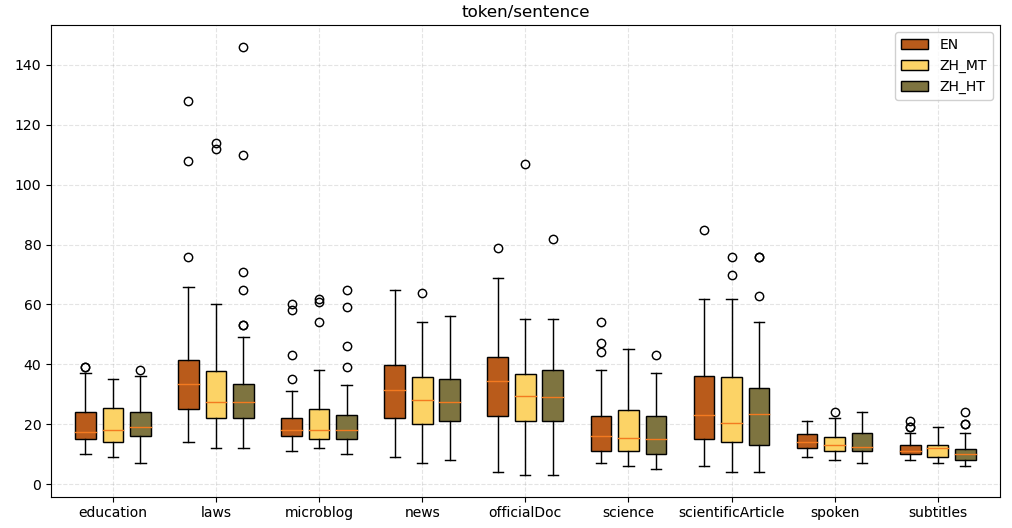}
\caption{Figure 3: Input of source.txt file and target.txt file (Example from Liu (2015))}
\end{figure}

\begin{figure}[H]
\centering
\includegraphics[scale = 0.47]{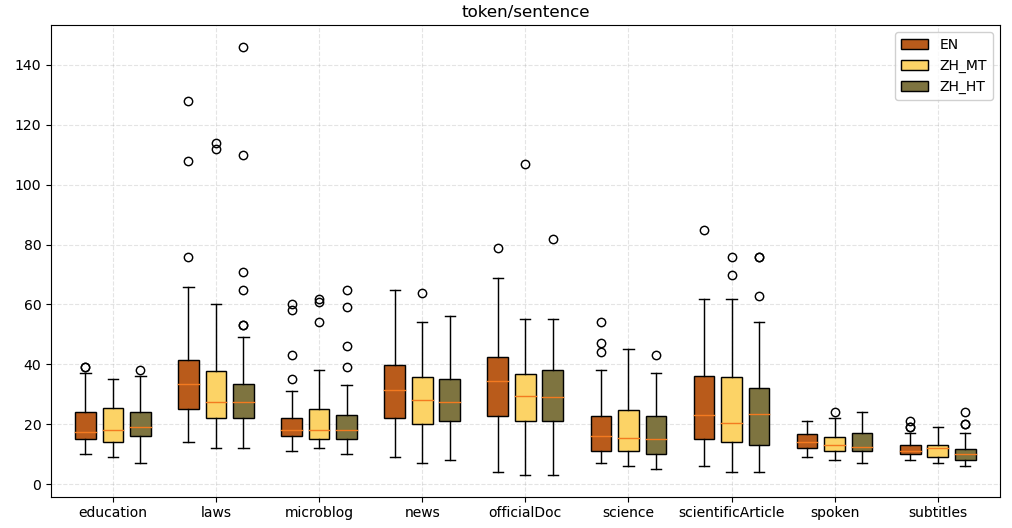}
\caption{Figure 4: Output of token index alignment (Example from Liu (2015))}
\end{figure}

When tokens of parallel sentences are aligned and turned into token index matches, both index alignment files in the format of .aln and tokenized parallel sentences in the .crp files are uploaded onto YAWAT. YAWAT is a tool for manipulating and visualizing word- and phrase-level parallel text alignments (Germann, 2008). This tool is a webpage-based platform, enabling manual word alignment and labeling such as labeling for translation relations or translation techniques. On the YAWAT operation webpage (see Figure 5) with two parts divided, the left side contains source sentences while the right is for target sentences. For word alignment between source and target texts, there is a matrix for each alignment, in which the upright side represents the source sentences and the horizontal is for target ones; in a matrix, there are many small blocks that are matches between the source and the target (see Figure 6). Once the word alignment is done for one sentence, the annotation for translation relations can be handled. The default translation technique is literal translation highlighted by the color yellow. If some other non-literal translation techniques are used, the labeling can be changed, and so can the colors with right clicks. There is a drop list of translation techniques with different colors for annotators’ convenience (see Figure 7). In the drop list on the right mouse click, the color yellow represents for literal translation, orange for equivalence, green for transposition, light blue for modulation, green for modulation + transposition, brown for generalization, red for particularization, pink for figurative, purple for lexical shift, light red for uncertain, and deep blue for translation error. 

\begin{figure}[H]
\centering
\includegraphics[scale = 0.3]{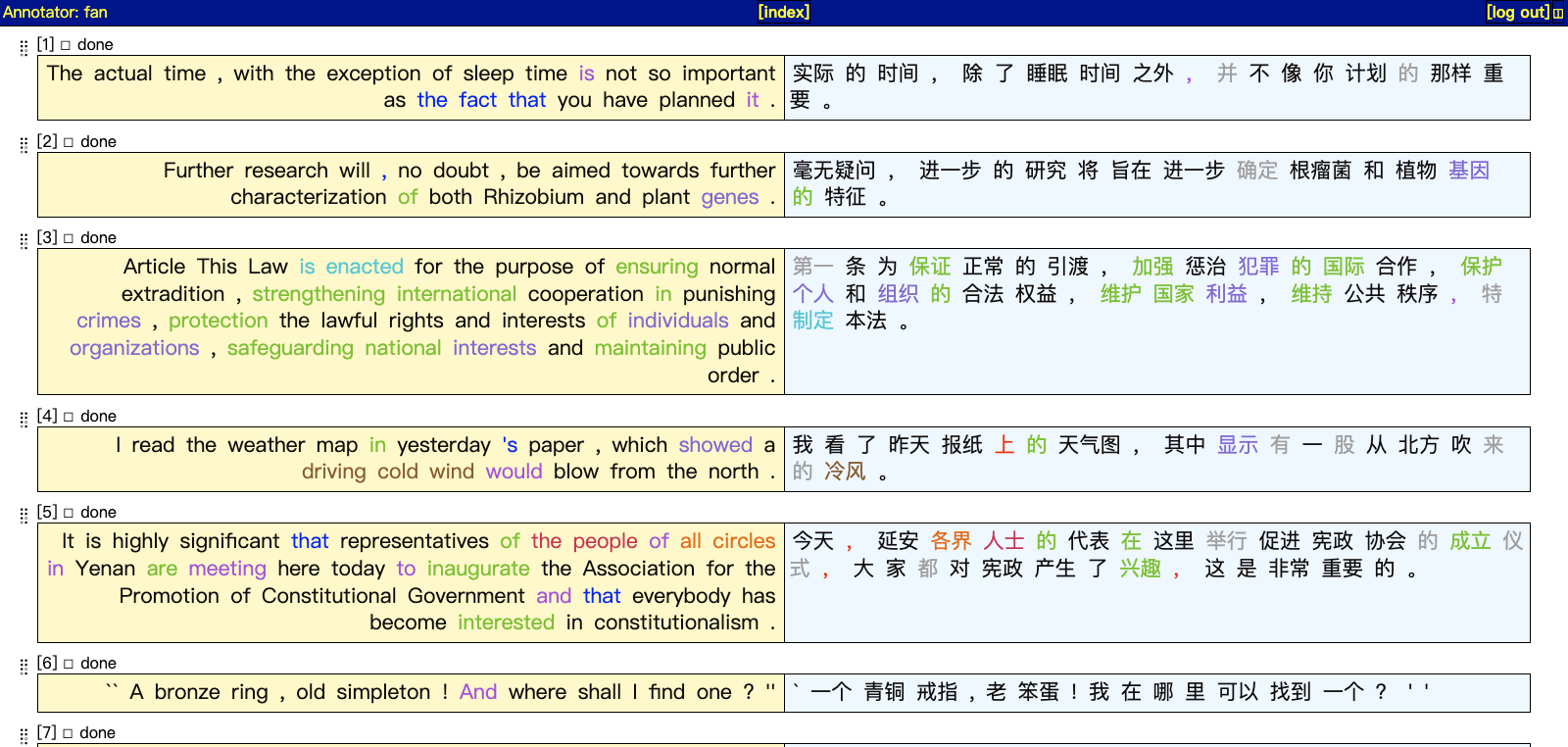}\\
\caption{Figure 5: The operation web page of YAWAT}
\end{figure}

\begin{figure}[H]
\centering
\includegraphics[scale = 0.3]{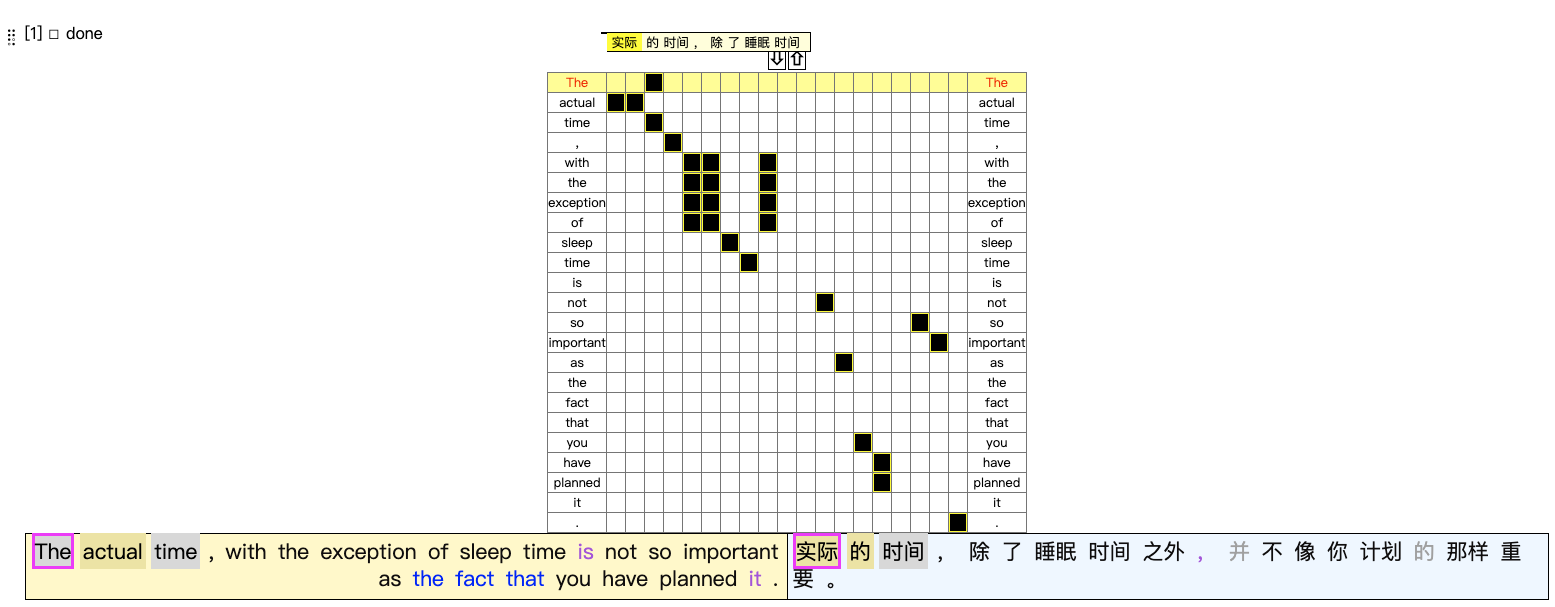}\\
\caption{Figure 6:  A matrix for word alignment, and the black block used for matching source token groups and the target token groups}
\end{figure}

\begin{figure}[H]
\centering
\includegraphics[scale = 0.3]{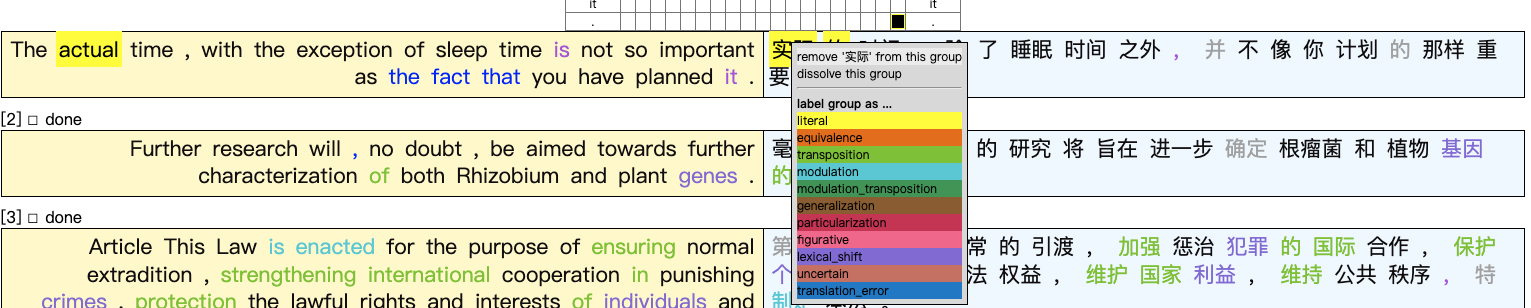}\\
\caption{Figure 7: The drop list of labeling for translation relations contains 11 different translation techniques with different colors}
\end{figure}

Inconsistency of translation relation annotation was found in the HT data due to the fact that there are three annotators to annotate HT data. Therefore, HT data annotation was modified to make the whole labeling consistent. 
\pagebreak
\subsection{Translation relations features}

The discussion and comparison of linguistics-related traits and components that lead to the use of non-literal translation procedures between HT and MT data. Literal translation, translation mistake, uncertainty, and non-type are among the 14 categories in translation relations that are not covered. The remaining 10 non-literal translations will be analyzed in terms of in terms of semantics and syntax in linguistics. \\

In general, the traits and factors that trigger the use of non-literal translations can be classified into semantics-related and syntax-related ones. For factors that involving semantic knowledge, there are equivalence which have 5 sub-types such as small changes of word meaning, fixed-expressions, etc., figurative translation, and generalization. The features are analyzed and classified through observations and some automatic semantic classification tools such as SpaCy’s named entity recognition. There are some non-literal translation techniques’ characteristics relates to syntactic knowledge, covering lexical shift, transposition, unaligned explicitation and reduction. The analyses for these traits are based on some language-independent metrics including POS and dependency relations. Besides, some techniques use a kind of hybrid method that combines semantic and syntactic knowledge for analyses and comparison, including modulation, modulation+transposition, and particularization. \\

Based on these fine-grained classified factors or features, HT and MT data are compared on more subtle levels to find the occasions of causing the use of translation relations. 
\pagebreak
\section{\uline{Discussions and findings}}
In this section, the use of translation relations in HT and MT data is analyzed and compared from various perspectives. Besides, there are linguistics-related factors that lead to the use of a certain translation technique, and the used of translation relations caused by these factors are also analyzed and compared between HT and MT data. 
\subsection{Aligned Data}
In aligned MT and HT corpora, there are 9314 and 9235 aligned pairs which constitute the smallest unit for alignment. These aligned pairs can continue to be annotated based on different translation techniques. Generally, pairs can be divided into the type of literal translation which is a kind of common translation technique, and the type of non-literal translation which contains some other translation techniques such as equivalence, modulation, etc. Among these aligned pairs, literal translation accounts for the largest part both for MT and HT corpora. In comparison of ratio of aligned pairs annotated with literal translation , MT data uses more literal translation than HT data by 24.23\%, which verifies that machine translation systems are more inclined to use literal translation. However, It is noticeable that not all translation output delivered by machine translation systems is based on literal translation; around 23\% of aligned pairs use other translation relations. Specific data on aligned pairs using literal and non-literal translation is shown in Table 4.

\begin{center}
\setlength{\tabcolsep}{7mm}
\begin{tabular}{@{}lcccc@{}} 
\toprule
\multirow{2}{*}{Literal/non-literal} & \multicolumn{2}{c}{No. of pairs aligned} & \multicolumn{2}{c}{Percentage} \\
\cmidrule(lr){2-3} \cmidrule(lr){4-5}
 & MT data & HT data & MT data & HT data \\
\midrule
Literal & 6221 & 5045 & 76.895\% & 63.508\% \\
Non-literal & 3093 & 4190 & 23.105\% & 36.492\% \\
\midrule
Total & 9314 & 9235 & 100\% & 100\% \\
\bottomrule
\end{tabular}
\end{center}
\begin{center} Table 4 Token number of each genre in source texts, machine translated texts, and human translated texts
\end{center} 

More details of aligned pairs annotated with different translation techniques can be seen in Table 5. Even though the total numbers of aligned pairs are different between MT and HT aligned corpora, which are caused by the different translations in the target language, there is no large difference between these two corpora, and the number of pairs can be seen as equivalent. However, on the basis of translation relations, there are distinct differences between these two aligned data sets. As the absolute total numbers are not the same, the percentage of relative numbers are compared. For the techniques of literal translation, lexical shift, and translation errors, MT data uses them more frequently than HT data, especially by using literal translation, 21\% more than HT data. As to translation error, the discrepancy is large reaching 63\%; however, the pair number of translation errors is small for evaluation. On the one hand, we can say that translation errors have more chances of occurring in MT data, while on the other hand, it is hard to evaluate what the real discrepancy is between MT and HT data on translation errors with small amount of data. There are some other types of non-literal translation techniques used more in HT data than in MT data, such as equivalence, particularization, explicitation, and reduction. For generalization, modulation, and transposition, there are smaller gap for techniques used between HT and MT data. Figurative techniques and modulation plus transposition are rarely employed in both HT and MT data, but they are still more used in HT data. 

\begin{center}
\setlength{\tabcolsep}{4mm}
\begin{tabular}{@{}lccccc@{}} 
\toprule
\multirow{2}{*}{Translation relations} & \multicolumn{2}{c}{No. of pairs aligned} & \multicolumn{2}{c}{Percentage} & \multirow{2}{*}{Discrepancy (MT data)} \\ 
\cmidrule(lr){2-3} \cmidrule(lr){4-5}
 & MT data & HT data & MT data & HT data &  \\
\midrule
equivalence & 164 & 326 & 2.137\% & 3.952\% & -45.926\% \\
figurative & 1 & 3 & 0.011\% & 0.043\% & -74.419\% \\
generalization & 32 & 49 & 0.376\% & 0.552\% & -31.884\% \\
lexical\_shift & 756 & 705 & 9.244\% & 8.652\% & 6.842\% \\
literal & 6221 & 5045 & 76.895\% & 63.508\% & 21.079\% \\
modulation & 95 & 128 & 1.299\% & 1.689\% & -23.091\% \\
modulation\_transposition & 18 & 22 & 0.204\% & 0.249\% & -18.072\% \\
particularization & 102 & 224 & 1.310\% & 3.064\% & -57.245\% \\
translation\_error & 21 & 14 & 0.258\% & 0.162\% & 59.259\% \\
transposition & 500 & 545 & 6.206\% & 6.724\% & -7.704\% \\
no\_type & 30 & 62 & 0.344\% & 0.845\% & -59.290\% \\
unaligned\_explicitation & 736 & 1220 & 7.902\% & 13.211\% & -40.184\% \\
unaligned\_reduction & 636 & 870 & 6.828\% & 9.421\% & -27.517\% \\
uncertain & 2 & 22 & 0.032\% & 0.282\% & -88.652\% \\
\midrule
Total & 9314 & 9235 & 100\% & 100\% & 0 \\
\bottomrule
\end{tabular}
\end{center}
\begin{center} Table 5 The number of aligned pairs using literal and non-literal translation
\end{center}

\subsubsection{Comparison based on tokens in the source texts}

The same source texts are shared by the aligned MT and HT corpora. For each token in the source texts, the application of translation relations is examined. In general, there are 10916 tokens in the source texts. 7623 and 6388 English tokens in MT and HT corpora are translated into Chinese using the literal translation approach respective. Table 6 provides more precise statistics.\\

The ratio of the number of literally translated tokens in a phrase is examined in accordance with various genres in order to compare the literally translated tokens in the source texts between the MT and HT data sets (see Figure 8). It is clear from the statistics that the ratio of literally translated tokens in the MT data is higher than that in the HT data. The education, microblog, scientific article, spoken language, and subtitles categories have the most disparities when utilizing literal translation.

\begin{center}
\setlength{\tabcolsep}{7mm}
\begin{tabular}{@{}lcc@{}} 
\toprule
English token number & MT data & HT data\\
\midrule
Literal token & 7623 & 6388 \\ 
Non-literal token & 3293 & 4528 \\ 
\midrule
Total & 10916 & 10916 \\ 
Percentage (literal) & 69.833\% & 58.520\% \\ 
\bottomrule
\end{tabular}
\end{center} 
\begin{center} Table 6 The number of literally and non-literally translated tokes used in source texts 
\end{center}

\begin{figure}
\centering
\includegraphics[scale = 0.4]{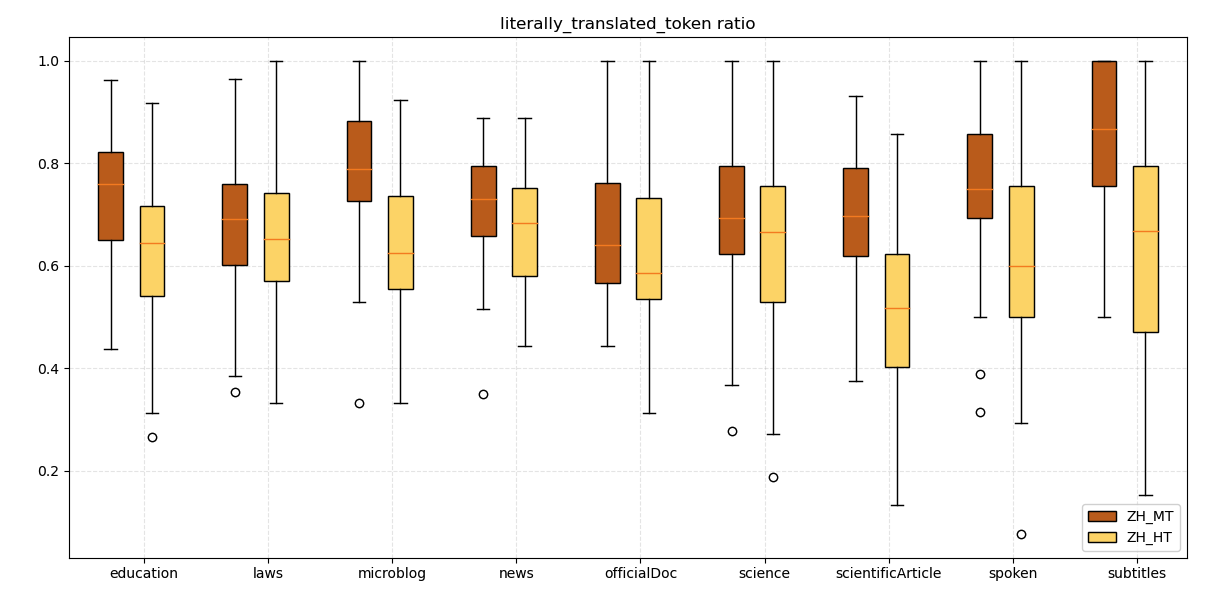}\\
\caption{Figure 8:  The proportion of literally translated token ratio in each sentence}
\end{figure}

\begin{CJK}{UTF8}{gbsn}The utilization of translation relations in MT data and HT data are compared using the edit distance to determine how different they are.  Since English tokens from the source texts are shared in both corpora, they each utilize a separate set of translation relations. For example, in the sentence “When my mother sees this rip in my new dress she will raise the roof .”, its translations are “当 我 母亲 看到 我 新 衣服 上 的 裂缝 ， 她 会 非常 生气 。” by human translators, and “当 我 母亲 看到 我 新 衣服 上 的 这 道 裂痕 时 ， 她 会 掀起 屋顶 。” by machine translation systems. As to “raise the roof” these three tokens, their translation in HT is “非常 生气” (very angry) by using generalization, while in MT data, they use literal translation with translation as “掀起 屋顶” (lift the roof). The difference of annotation can be seen in Figure 9 and Figure 10. \end{CJK}

\begin{figure}[H]
\centering
\includegraphics[scale = 0.3]{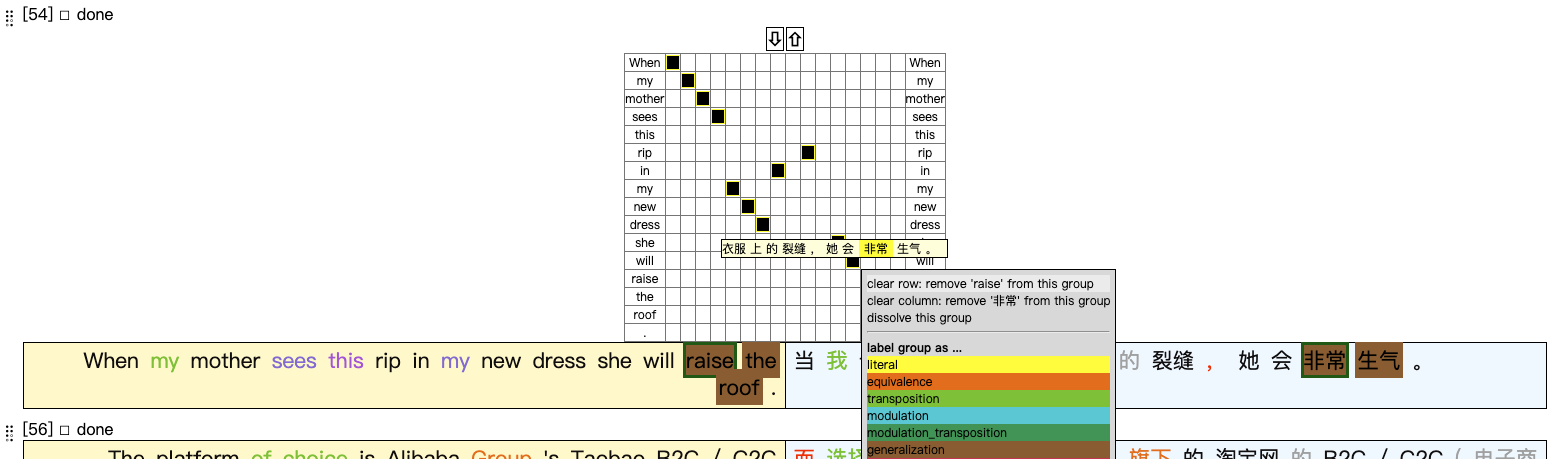}\\
\begin{CJK}{UTF8}{gbsn}\caption{Figure 9: "raise the roof" is translated into "非常生气" (very angry) by using "generalization" (brown part in the annotation)}\end{CJK}
\end{figure}

\begin{figure}[H]
\centering
\includegraphics[scale = 0.3]{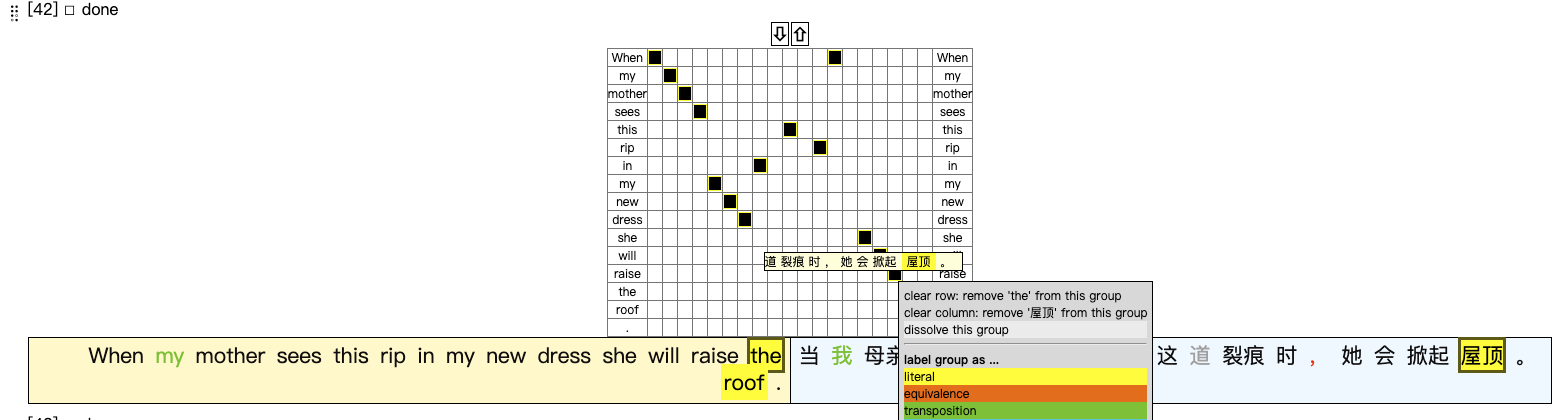}\\
\begin{CJK}{UTF8}{gbsn}\caption{Figure 10: "raise the roof" is literally translated into "掀起屋顶" (the roof)(yellow part in the annotation)}\end{CJK}
\end{figure}

In this case, the core idea of Levenshtein distance is used to examine how different translations are used in HT and MT data, and there is an example for the calculation of edit distance. In this case, the edit distance of translation relations instead of characters is calculated in this example sentence, and the edit distance is 3 since "raise","the" and "roof" are translated with "generalization" respectively in HT data while they are literally translated in MT data (see Table 7). The collections of edit distances of sentences in each genre are shown in Figure 11. The edit distance of each sentence in subtitle is noticeably greater than that of sentences from other genres, demonstrating the similarity in how translation is used in spoken and scientific publications.

\begin{center}
\setlength{\tabcolsep}{1mm}
\begin{tabular}{@{}cccccccccccccccc@{}} 
\toprule
data & When & my & mother & sees & this & rip & in & my & new & dress & she & will & raise & the & roof\\
\midrule
HT & literal & literal & literal & literal & literal & literal & literal & literal & literal & literal & literal & literal & literal & literal & literal \\ 
MT & literal & literal & literal & literal & literal & literal & literal & literal & literal & literal & literal & literal & \makecell{general-\\ization} & \makecell{general-\\ization} & \makecell{general-\\ization} \\ 
edit & 0 & 0 & 0 & 0 & 0 & 0 & 0 & 0 & 0 & 0 & 0 & 0 & 1 & 1 & 1 \\ 
\bottomrule
\end{tabular}
\end{center}

\begin{center} Table 7 the edit distance calculation of translation relations in an example sentence with the edit distance of 3
\end{center} 

\begin{figure}[H]
\centering
\includegraphics[scale = 0.55]{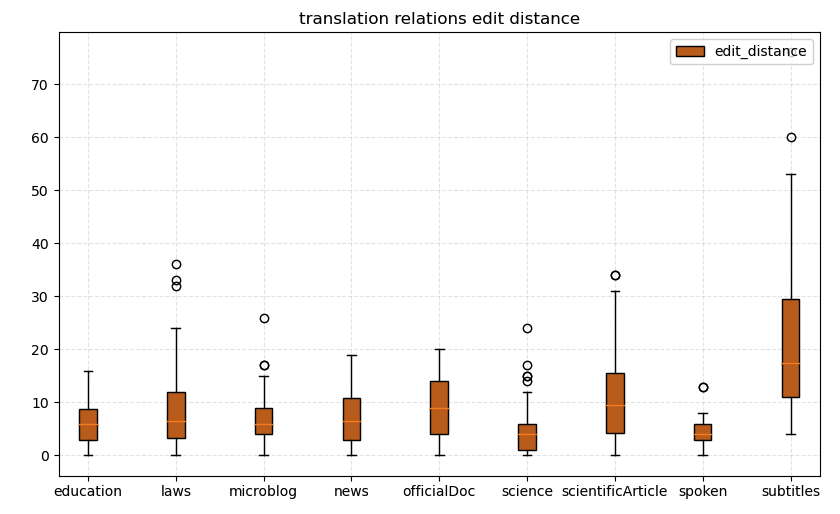}\\
\caption{Figure 11: Edit distance of translation relations in each sentence in genres}
\end{figure}

\subsubsection{Comparison of translation relations in genres}
Literal and non-literal translation techniques are analyzed based on different genres in Table 8. It is clear that non-literal translation procedures are used more frequently in HT data than in MT data; in fact, over half of HT data have done so, compared to just one-third of MT data. The difference in subtitle is 98\%, and it is also noticeable in spoken, microblog, and scientific articles. Besides, there is a significant disparity between the usage of non-literal translation in official documents, legislation, and science, which demonstrates that formal, rigid writings are more prone to employ literal translations.

\begin{center}
\setlength{\tabcolsep}{4mm}
\begin{tabular}{@{}lccccccc@{}} 
\toprule
\multirow{2}{*}{Genre} & \multicolumn{2}{c}{MT data} & \multicolumn{2}{c}{HT data} & \multicolumn{3}{c}{Percentage (non-literal)} \\
\cmidrule(lr){2-3} \cmidrule(lr){4-5} \cmidrule(lr){6-8}
 & Literal & Non-literal & Literal & Non-literal & MT data & HT data & Discrepancy \\
\midrule
education & 613 & 273 & 496 & 382 & 30.813\% & 43.508\% & 41.202\% \\
laws & 941 & 563 & 820 & 636 & 37.434\% & 43.681\% & 16.690\% \\
microblog & 747 & 282 & 562 & 464 & 27.405\% & 45.224\% & 65.020\% \\
news & 906 & 411 & 805 & 528 & 31.207\% & 39.610\% & 26.925\% \\
officialDoc & 877 & 620 & 783 & 654 & 41.416\% & 45.511\% & 9.888\% \\
science & 545 & 288 & 487 & 353 & 34.574\% & 42.024\% & 21.548\% \\
scientificArticle & 748 & 365 & 475 & 695 & 32.794\% & 59.402\% & 81.135\% \\
spoken & 421 & 179 & 341 & 283 & 29.833\% & 45.353\% & 52.020\% \\
subtitles & 423 & 112 & 276 & 195 & 20.935\% & 41.401\% & 97.765\% \\
\midrule
Total & 6221 & 3093 & 5045 & 4190 & 33.208\% & 45.371\% & 36.626\% \\
\bottomrule
\end{tabular}
\end{center}
\begin{center} Table 8 Literal and non-literal translation techniques used in different genres
\end{center} 

Non-type and ambiguous translation relations, out of the 14 relations, won't be explored in this study. Instead, more information and trends of genres employing the same translation techniques are evaluated (see Figure 12).

\begin{figure}[H]
\centering
\includegraphics[scale = 0.35]{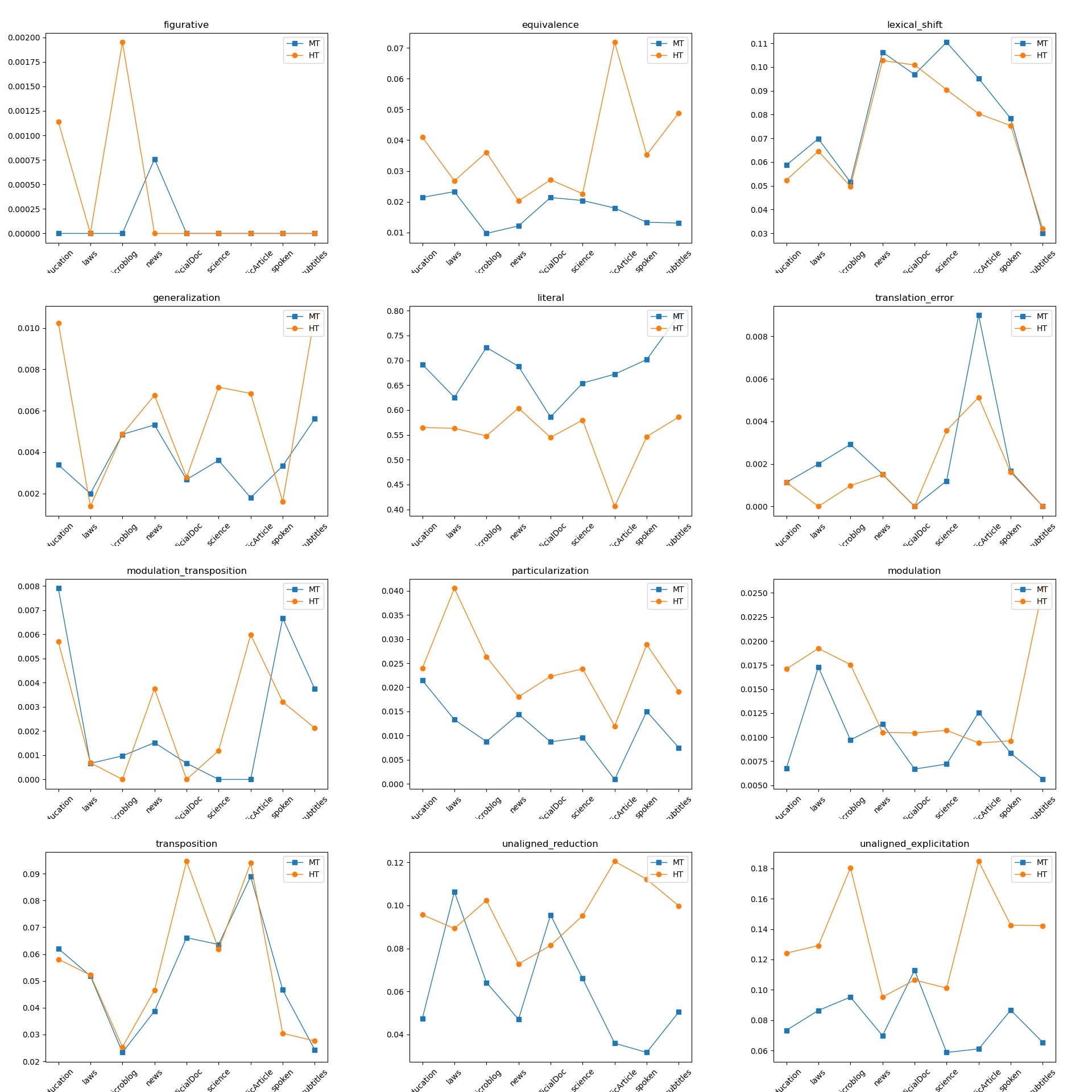}\\
\caption{Figure 12: Percentages of use of translation relations in different genres}
\end{figure}

(1) Equivalence\\
The technique of equivalence is used in over 2\% of all HT data, with scientific article texts using it the most frequently (at 7\%), followed by subtitles at over 5\%. In contrast, the total application of equivalence for MT data is between 1.5\% and 2.3\%, with the exception of microblogs, whose share is much lower than the flat trend of less than 1\%.

(2) Figurative\\
Figuration contains metaphors and idioms, making it a challenging strategy for both human translators and machine translation systems. Personification is occasionally employed to cope with certain linguistic situations. Only 4 pairings of the metaphorical approach are used in the HT data in these two matched data sets, and only 1 pair is used in the MT data. Even though there is not a lot of data to analyze, it demonstrates that human translators are more mindful when using it.

(3) Generalization\\
In HT data, the education and subtitle genres employ generalization the most; the spoken and legal genres use it the least, much less than in MT data. Using the generalization approach, two sets of data exhibit the same level.

(4) Lexical shift\\
Lexical shift is utilized somewhat more in MT data than HT data. Two aligned corpora exhibit extremely comparable patterns, with spoken texts, subtitles, officialDoc, science, scientificArticle, and microblog employing lexical shift significantly more frequently than laws, education, microblog, and officialDoc.

(5) Literal translation\\
For all genres in the MT data set, the percentage of literal translation exceeds 60\%, and for subtitles, it even approaches 80\%. With the exception of scientificArticle, where the ratio is lower than 40\% for HT-aligned corpora, the use of literal translation stabilizes between 55\% and 60\%.

(6) Modulation\\
In HT data, modulation is applied 1\% more often to texts in the legal, scholarly, microblog, and subtitle genres than to texts in the other genres. In contrast to HT data, the usage of modulation in educational texts and subtitles is least for MT data.

(7) Modulation + transposition\\
Both in HT and MT corpora, this combination of translation methods is uncommon. With the exception of a scientific article, both sets of data show comparable tendencies, with HT data showing a higher adoption of the combining approach than MT data.

(8) Particularization\\
There is a distinct line that separates the MT and HT corpora: HT data is distributed on the upper line, while MT data is below it. In legal writings, microblogs, and scientific papers, particularization is employed more cautiously and conservatively in MT data.

(9) Transposition\\
The same trends and comparable data in many genres are shared by the MT and HT databases. However, there are two exceptions: the first is that MT utilizes transposition less frequently in official papers than HT, and the second is that HT uses it less frequently in spoken speech than MT.

(10) Translation error\\
Across all genres, the MT data set often contains more translation mistakes than the HT data. The proportion of translation mistakes in MT and HT datasets highest in scientific articles.

(11) Unaligned-explicitation \\
Explicitation only occurs in the source materials, and certain additional details are given to the translation to make it seem more natural. More explicitation is used in the HT data set than the MT data. Less of this technique is used in the HT data set's news, government papers, and science texts.

(12) Unaligned-reduction\\
Reduction is applied to tokens in the original texts that are not translated. Microblogs and scientific articles used reduction more frequently in aligned HT corpora, while MT data used it more frequently in legal and official texts.

Overall, HT data uses more non-literal translation techniques than MT data. Besides, the developing trends of two types of data in most of genres are similar except some slight points are out of steps. Two lines which are nearly in the same places when the translation techniques of lexical shift and transposition are used, which illustrates that GNMT has almost reached human parity when using these two translation techniques which are on syntactic levels. 

\subsection{Features of translation relations}
Different translation techniques each have unique characteristics and linguistic causes that lead to the usage of a certain type of translation technique. This section compares this MT and HT data in terms of such characteristics and elements that lead to a certain translation technique.

\subsubsection{Equivalence}

Equivalence is generally applied twice as often in HT as it is in MT. However, not all matched pairings with equivalence annotations fall under the same linguistic classification. With observations and summarizing, there are five linguistic sub-categories covering slight semantic changes, named entities, phrase (fixed-expression), adjective translation, and refined translations which may be used to classify all aligned pairings using equivalence. As MT and HT share the same definitions, the examples chosen for these five kinds are taken from either MT or HT data. More numerical details can be seen in Table 9. More information about these five types are examplified. 

\begin{center}
\setlength{\tabcolsep}{7mm}
\begin{tabular}{@{}lcc@{}} 
\toprule
Sub-categories of equivalence & MT data aligned pairs & HT data aligned pairs \\
\midrule
Slight semantic changes & 78 & 113 \\
Named entity & 40 & 83 \\
Adjective & 14 & 14 \\
Refined words & 4 & 63 \\
\midrule
Total & 136 & 273 \\
\bottomrule
\end{tabular}
\end{center}
\begin{center} Table 9 Five types of equivalence used in aligned pairs
\end{center}

(1) slight semantic changes: Unlike literal translation, aligned translations of source texts have somewhat different meanings in the destination texts. However, this semantic similarity between aligned pairings is less than that of synonyms and more than that of completely distinct terms. 

Example 1: \\
Source sentence: Experience is not interesting till it begins to repeat itself , in fact , till it does that , it hardly is experience.\\
\begin{CJK}{UTF8}{gbsn}Target sentence: 经验 是 没有 意义 的 ， 直到 它 开始 重复 自己 ， 事实上 ， 直到 它 这样 做 ， 它 很 难 成为 经验 。\end{CJK}

\begin{CJK}{UTF8}{gbsn}The translation of the word "interesting" from the original text is “有意义的” (meaningful, significant). Despite the discrepancies between the two terms, it is not considered a translation mistake because it makes sense for the entire statement. In both HT and MT data, the category of "slight changes in semantics" has the biggest proportion in the application of equivalence, as shown in Table 7. However, the amount of data in HT corpora that falls under this category is 1.5 times more than that in MT corpora.\end{CJK}

(2) Named entities: Named entities can be identified by a proper name, such as a person, place, business, product, etc. 

Example 2: \\
Source sentence: ( afp , washington ) the world health organization said that in china and india , the two countries whose combined tb cases account for one third of the world 's total , great progress has been made in the prevention and treatment of this infectious disease .\\
\begin{CJK}{UTF8}{gbsn}Target sentence: （ 法新社 华盛顿 ） 世界 卫生 组织 说 ， 在 中国 和 印度 这 两 个 结核病 病例 合并 占 世界 总数 三分之一 的 国家 ， 在 防治 这 一 传染病 方面 取得 了 很 大 进展 。\end{CJK}

"TB" stands for tuberculosis in the original language. Equivalence is applied to deal with such named entities. Named entities which result in the application of equivalence The second most common scenario from Table 7. The fact that HT data recognizes and translates twice as many name entities as MT data shows that NMT is unable to recognize some name entities, such as "OO" (the acronym for "object-oriented") in MT data.

(3) phrase (fixed-expression): Some sentences have their own fixed translations that cannot make sense by a literal translation.

Example 3: \\
Source sentence: But in one respect I have succeeded as gloriously as anyone who 's ever lived : I 've loved another with all my heart and soul ;\\
\begin{CJK}{UTF8}{gbsn}Target sentence: 但 在 某 一 方面 ， 我 的 成功 与 以往 任何人 一样 光荣 ： 我 全心全意 地 爱 着 另 一个 人 ；\end{CJK}

\begin{CJK}{UTF8}{gbsn}If "with all my heart and soul" in the original text is translated literally, the result is “用我的真心和灵魂” which does not match the Chinese idiomatic expressions. Instead, the comparable statement in Chinese is represented by a four-character word “全心全意” (whole-heartedly). For fixed phrases, there are still twice as many translations in HT than there are in MT data.\end{CJK}

(4) Adjective translation: In order to make adjectives seem more natural and catchy in Chinese, adjectives are frequently translated into "four-character" terms.

Example 4: \\
Source sentence: he said : " she is just like the girl next door , extremely amiable and also with an inspiring life story . ”\\
\begin{CJK}{UTF8}{gbsn}Target sentence: 他 说 ： “ 她 就 像 邻 家 女孩 一样 ， 非常 和蔼可亲 ， 也 有 励志 的 人生 故事 。 ”\end{CJK}

Example 5: \\
Source sentence: Now , in research that will be embraced by millions of bleary - eyed Britons\\
\begin{CJK}{UTF8}{gbsn}Target sentence: 现在 ， 在 一 项 将 被 数百万 睡眼惺忪 的 英国 人 所 接受 的 研究 中 \end{CJK}

Translations of "amiable" and "bleary-eyed" into "four-character" terms make them more consistent with Chinese idioms and styles. The performance in HT data is equivalent to that of MT data.

(5) Refined translations: While it is permissible to translate words or texts exactly as they appear in the source language, the translators would improve the translation by taking the texts' styles into account in order to make the translations better fit the target language's contexts and expressions.

Example 6:\\
Source sentence: A wind had cleared the mist , the autumn leaves were rustling and the stars were shining .\\
\begin{CJK}{UTF8}{gbsn}Target sentence: 一阵 风 吹 散 了 雾霭 ， 秋叶 沙沙 作响 ， 繁星 闪烁 。\end{CJK}

Example 6 employs literary terms to translate a text that portrays natural scenery and illustrates an aesthetic idea. The difference between the two sets of data on refined terms is enormous. Only four pairs in the MT data employ the revised translation, compared to 64 pairs in the HT data, which clearly shows how little capacity the GNMT system has to refine the translation based on contextual information.

To summarize, equivalence is a typical translation technique involves semantic knowledge on appropriate word selection to conform the context and styles in the target languages. It is evident that GNMT does perform well in this aspect.

\subsubsection{Figurative translation}

\begin{CJK}{UTF8}{gbsn}The goal of figurative translation is to explain a non-metaphor or non-fixed term using an idiom or metaphor; personification is also occasionally used. It occasionally takes contextual information into account. For instance, the word "simple" can be rendered as ‘柴米油盐’ (an idiom that means basic life necessities). In all aligned pairs, there are only four aligned pairs using figurative translation in HT and one in MT data, which reflects two issues: first, it is very difficult to be utilized in both human translation and machine translation, and second, the amount of data that is now available is extremely minimal, making it impossible to do analysis for this kind of technique. \end{CJK}

\subsubsection{Generalization}

\begin{CJK}{UTF8}{gbsn}When the technique of generation is examined, it mostly depends on two ways. The sentences in the brief names are chopped into multiple sections in the original materials, and the translation only uses one or a few of those portions. For example, in the phrase “anyone who is” is translated into “任何人” (anyone) instead of “任何人，他是”, and “refrain from” is translated into “不” (not) rather than “不让”. The second technique involves utilizing hyperonyms in the translated text, which implies that the phrase is used more general than in the original text. For example, “troops” is translated into “人” (people), and “travelling back” is translated into “返回” (returning back) rather than “旅行回来”. \end{CJK}

\begin{center}
\setlength{\tabcolsep}{7mm}
\begin{tabular}{@{}lcc@{}} 
\toprule
Sub-categories of generalization & MT data & HT data \\
\midrule
Short names & 13 & 4 \\
Hyperonyms & 19 & 45 \\
\midrule
Total & 32 & 49 \\
\bottomrule
\end{tabular}
\end{center}
\begin{center} Table 10 Two methods of generalization used in aligned pairs
\end{center} 

The application of short names and hyperonyms is calculated separately for HT and MT data; the particular data are shown in Table 10. Although the amount of data in both data sets is comparably small, the usage of hyperonyms in HT corpora is far higher than that in MT data, demonstrating that the NMT system does not have the same potential for generalization as human translators.

\subsubsection{Lexical shift}

In both aligned corpora, the lexical shift is a frequently used approach which has the typical attribute of syntax.  When verbs in phrases include traces of the present tense or the past tense, or when nouns are plural, lexical shift is used. According to Table 11, the number of tense and plural nouns that trigger lexical shift in MT data is a little higher than it is in HT data.

\begin{center}
\setlength{\tabcolsep}{7mm}
\begin{tabular}{@{}lcc@{}} 
\toprule
Sub-categories of lexical\_shift & MT data & HT data \\
\midrule
Plural nouns & 455 & 432 \\
Tense & 301 & 273 \\
\midrule
Total & 756 & 705 \\
\bottomrule
\end{tabular}
\end{center}
\begin{center} Table 11 Two cases of lexical shift used in aligned pairs
\end{center} 

\subsubsection{Modulation}

\begin{CJK}{UTF8}{gbsn}Irony and the switch from passive to active voice are the two distinctive characteristics that cause modulation to be used. In English, the passive voice is widely employed to show objectivity by combining "be+past tense of verb". However, passive voice is uncommon in Chinese; instead, active voice is frequently used. The conversion of passive speech to active voice is therefore one characteristic of modulation. Irony is also a characteristic of this approach. It's a figure of speech that emphasizes a point by saying the exact opposite of what is meant; the words' intended meaning is the complete reverse of their usual meaning. For example. “difficult” is translated into “不容易” (not easy) by using irony of modulation. The conversion from passive to active voice is the most frequently used category in modulation, and GNMT system can achieve the same performance with that produced by human translators. In the meanwhile, irony is utilized less frequently in both HT and MT data, but HT data in this category is twice as common as MT data.\end{CJK}

According to the sub-categories, modulation is a method that combining  both syntactic and semantic knowledge since the voice conversion belongs to syntactic level while irony to semantic level. That GNMT's performance is good on the former factor but bad on the latter once asserts that GNMT's performance on syntax is better that on semantics.

\begin{center}
\setlength{\tabcolsep}{7mm}
\begin{tabular}{@{}lcc@{}} 
\toprule
Sub-categories of modulation & MT data & HT data \\
\midrule
Passive voice to active voice & 72 & 70 \\
Irony & 11 & 23 \\
Others & 12 & 35 \\
\midrule
Total & 95 & 128 \\
\bottomrule
\end{tabular}
\end{center}
\begin{center} Table 12 Two cases of lexical shift used in aligned pairs
\end{center}

\subsubsection{Modulation + Transposition}
\begin{CJK}{UTF8}{gbsn}The most common use of “modulation+transposition” is the translation of propositional phrases. While the meaning remains the same, the manner of syntactic expressions changes, the proposition will be transferred to another POS labeling, something verbs. For example, “in” is translated into “用”(use), where both meaning and POS have changed. Such case in HT data is nearly twice than that in MT data according to Table 13. \end{CJK}

\begin{center}
\setlength{\tabcolsep}{7mm}
\begin{tabular}{@{}lcc@{}} 
\toprule
Sub-categories of modulation + transposition & MT data & HT data \\
\midrule
Proposition & 11 & 18 \\
Others & 7 & 4 \\
\midrule
Total & 18 & 22 \\
\bottomrule
\end{tabular}
\end{center}
\begin{center} Table 13 Two cases of lexical shift used in aligned pairs
\end{center}

\subsubsection{Particularization}

\begin{CJK}{UTF8}{gbsn}To determine how information from the original text is translated by applying particularization, part-of-speech tagging is employed. The computation of data processing shows that pronouns, nouns, verbs, adjectives, and adverbs are the five categories of POS terms that are most likely to be translated by using the technique of particularization. For example, several “it” are contained in one sentence, and if all “it” are translated into “它” (it), the translation into Chinese would be terrible since numerous instances of “它” (it) do not adhere to Chinese expressions. Each POS type in HT data typically employs twice as much method as it does in MT data. It requires contextual information to name a single pronoun or term from the source texts in the translation, demonstrating how inadequate GNMT's understanding of contexts is.\end{CJK}

\begin{center}
\setlength{\tabcolsep}{7mm}
\begin{tabular}{@{}lcc@{}} 
\toprule
POS-based sub-categories of particularization & MT data & HT data \\
\midrule
Pronoun & 23 & 47 \\
Noun & 41 & 87 \\
Verb & 23 & 60 \\
Adv., Adj. & 15 & 30 \\
\midrule
Total & 102 & 224 \\
\bottomrule
\end{tabular}
\end{center}
\begin{center} Table 14 POS-categorized particularization
\end{center}

\subsubsection{Transposition}

\begin{CJK}{UTF8}{gbsn}Between the source and destination languages, POS labeling is mutually transferred during transposition. The most common POS transfers from the most recent MT and HT data sets are shown in Table 15. The most frequent POS transfers are ADP—PART, ADJ—NOUN, and NOUN—VERB. The most ADP—PART pairs are used because there are many instances of "of" in phrases and "that/which" in attributive clauses, both of which are translated into “的” (’s) in Chinese. For ADJ—NOUN, it is rare to see that “…的” in Chinese, and adjectives sometimes are converted into “generic verb” + “a noun”. For example, “homophonic”, an adjective is translated into “谐音”, a noun.  Translation from nouns to verbs is also a common see, as the Chinese language uses more verbs than nouns (Choi et al., 1995; Kim et al., 2000; Ogura, 2001; Tardif, 1996; Imai et al., 2008).  \end{CJK}

\setlength{\tabcolsep}{7mm}{
\begin{center}
\begin{tabular}{@{}lcc@{}} 
\toprule
POS-transfer-based sub-categories of transposition & MT data & HT data \\
\midrule
ADP---PART & 120 & 113 \\
ADJ---NOUN & 69 & 79 \\
ADP---NOUN & 18 & 31 \\
ADJ---PROPN & 18 & 11 \\
PRON---PART & 17 & 17 \\
NOUN---VERB & 71 & 76 \\
ADJ---PART & 10 & 6 \\
ADP---VERB & 10 & 18 \\
ADJ---VERB & 11 & 14 \\
VERB---NOUN & 11 & 8 \\
ADJ---ADV & 7 & 7 \\
DET---PART & 6 & 5 \\
Other & 132 & 160 \\
\midrule
Total & 500 & 545 \\
\bottomrule
\end{tabular}
\end{center}

\begin{center} Table 15 POS transfers in transposition
\end{center} 

The transfer of word POS is totally one pure aspect of syntax. HT data only performs slight better than MT data when transposition is used, demonstrating that GNMT can perform well when transposition is used.

\subsubsection{Unaligned-explicitation}

\begin{CJK}{UTF8}{gbsn}Explicitation or additional information added in translation to make the output more natural is used in the source texts, and the additional information in the translation cannot be found in its corresponding source texts. Overall, in human translation, there is more additional information than MT data, showing that more consciousness is put into translation processes. The translation additionally added in the target texts is analyzed through POS tagging (see Table 16), and the most frequent occurrences are PART, VERB, NOUN and NUM of Chinese tokens. They can represent function words such as “了” and “就”, and number words like “一 头”,  “一 句” which are not used in English. \end{CJK}

\begin{center}
\setlength{\tabcolsep}{7mm}
\begin{tabular}{@{}lcc@{}} 
\toprule
POS-transfer-based sub-categories of unaligned-explicitation & MT data & HT data \\
\midrule
ADV & 71 & 148 \\
PART & 253 & 223 \\
VERB & 112 & 175 \\
NUM & 97 & 143 \\
NOUN & 109 & 262 \\
PRON & 16 & 17 \\
X & 18 & 11 \\
CCONJ & 7 & 17 \\
ADP & 26 & 46 \\
PUNCT & 11 & 53 \\
ADJ & 7 & 21 \\
DET & 3 & 23 \\
PROPN & 6 & 21 \\
SCONJ & 0 & 3 \\
Others & 0 & 37 \\
\midrule
Total & 736 & 1220 \\
\bottomrule
\end{tabular}
\end{center}
\begin{center} Table 16 Top added POS tagging in Unaligned-explicitation
\end{center} 

The tokens that do not have their counterpart in the sources texts are additionally translated in the target texts, and in this case, apart from analyzing POS of these additional tokens, it is similarly important to know what kind of dependency relations that these tokens are. Most dependency relations of explicitation in HT data are twice of those in MT data, among which punct, compound:nn, dobj, and nsubj in MT data are the worst. However, the dependency of mark in MT data is more than in HT data. It shows that GNMT is not apt to add additional words into translation which can make translation more natural. In this case, both POS and dependency relations are detected, which can give a direction for the follow-up work of optimizaing NMT systems on expliciation. 

\begin{center}
\setlength{\tabcolsep}{7mm}
\begin{tabular}{@{}lcc@{}} 
\toprule
DEP-transfer-based sub-categories of unaligned-explicitation & MT data & HT data \\
\midrule
case & 140 & 155 \\
advmod & 69 & 153 \\
dep & 46 & 95 \\
compound:nn & 22 & 94 \\
mark:clf & 64 & 89 \\
dobj & 30 & 77 \\
mark & 97 & 67 \\
nsubj & 19 & 56 \\
punct & 9 & 52 \\
conj & 23 & 48 \\
nmod:prep & 22 & 36 \\
acl & 15 & 32 \\
amod & 6 & 30 \\
ROOT & 19 & 29 \\
others & 155 & 207 \\
\midrule
Total & 736 & 1220 \\
\bottomrule
\end{tabular}
\end{center}
\begin{center} Table 17 Top added DEP tagging in Unaligned-explicitation
\end{center}

\subsubsection{Unaligned-reduction}

Some words or phrases from the original texts are dropped in the translated texts. HT data drops more English words in the source texts than MT data. The POS of terms that are most unlikely to be translated include ADP, CCONJ, AUX, PART, and DET, which are function words in English but are not utilized in Chinese, according to Table 18. We can see that HT data removes both function terms and semantic words more than MT data when comparing MT and HT disparities in abandoning English words in Chinese translation.

\begin{center}
\setlength{\tabcolsep}{7mm}
\begin{tabular}{@{}lcc@{}} 
\toprule
POS-based sub-categories of unaligned-reduction & MT data & HT data \\
\midrule
AUX & 60 & 72 \\
PRON & 46 & 108 \\
ADP & 222 & 244 \\
VERB & 26 & 40 \\
PART & 78 & 45 \\
CCONJ & 56 & 77 \\
PUNCT & 16 & 16 \\
ADJ & 20 & 36 \\
NOUN & 37 & 67 \\
SCONJ & 9 & 18 \\
DET & 36 & 40 \\
PROPN & 14 & 11 \\
X & 0 & 1 \\
NUM & 2 & 5 \\
INTJ & 2 & 2 \\
More than one token & 0 & 52 \\
\midrule
Total & 636 & 870 \\
\bottomrule
\end{tabular}
\end{center}
\begin{center} Table 18 Top dropped POS tagging in Unaligned-reduction
\end{center} 

Analysis is done on the dependency relations between English tokens that are not translated in the source texts. In general, HT data does not translate more of these tokens than MT data. In both MT and HT data, the dependency relation prep (prepositional modifiers) has the highest number of untranslated source tokens. However, in this dependence relation, the HT data has slightly more untranslated tokens than the MT data does. This little difference also affects cc, det, poss, ROOT, etc. In addition, several dependency relations, such as nsubj, advmod, pobj, and amod, demonstrate that the number of untranslated tokens in MT data is significantly lower than that of HT data. The smaller and greater disparities demonstrate that GNMT's translation performance varies and has space for improvement. However, despite the fact that GNMT's performance is generally inferior than that of human translators in many areas, the kind of aux in MT data is bigger than in HT data. The data can reach a conclusion that GNMT is more inclined to remain the target texts and make them translated into target language as much as possible. POS and dependencies of reduction in HT data are analyzed so as to make a guideline for the next step improvement for NMT system on the use of reduction.

\begin{center}
\setlength{\tabcolsep}{7mm}
\begin{tabular}{@{}lcc@{}} 
\toprule
DEP-transfer-based sub-categories of unaligned-reduction & MT data & HT data \\
\midrule
aux & 97 & 75 \\
nsubj & 19 & 71 \\
det & 35 & 57 \\
advmod & 13 & 52 \\
pobj & 21 & 50 \\
amod & 19 & 40 \\
poss & 22 & 36 \\
ROOT & 23 & 33 \\
punct & 15 & 22 \\
dobj & 13 & 21 \\
others & 112 & 80 \\
\midrule
Total & 636 & 870 \\
\bottomrule
\end{tabular}
\end{center}
\begin{center} Table 19 Top dropped DEP tagging in Unaligned-reduction
\end{center}


\pagebreak
\section{\uline{Conclusions and future work}}
\subsection{Research questions}
Translationese is pervasive in both human translation and machine translation. Studies and efforts are being made to quantify translationese using certain metrics, bring it down to human parity, and even to the level of non-translated texts in target languages. Translation relations, a technique for reducing translationese to make translation accurate, fluent, faithful, and natural, is employed in this dissertation's work as a metric to quantify translationese in translations from English to Chinese generated by both human translators and Google Translate. Translation relations show the relationships between various translation techniques, including literal and non-literal translation techniques as well as a few additional types of translation phenomena. In this study, both MT and HT are annotated with various translation techniques for comparison. Specifically, the aligned MT and HT corpora are compared on three different levels: the general differences between translation relations used in HT and MT are the first set of comparisons; the comparison of non-literal translation techniques used in HT and MT data is also focused; the features and factors that lead to the use of one specific non-literal translation technique in the MT and HT parallel corpora are compared. Comparisons highlight the differences between MT and HT capacities for application of translation relations so as to close this gap and improve machine translation in terms of using translation relations in the follow-up work.

\subsection{Data}
The study makes use of two parallel corpora. The first parallel corpus covers the English texts and its translation in Chinese is performed by human translators. The second parallel corpus covers the English texts shared with HT data as the source texts and its translation by GoogleNMT system (Google Translate). These two corpora are compared with HT data as a benchmark in order to identify the under-utilization of NMT translation from the standpoint of translation relations.
\subsection{Results}
The overall conclusion is that HT data performs better than MT data in terms of translation relations since MT data relies more heavily on literal translation (77\%) than HT data (64\%), which can be seen as a hint to have more translationese. In particular, lexical shift, transposition, and unaligned-reduction translation procedures with absolute proportion disparities of 6.8\%, 7.7\%, and 28.63\% respectively, indicate that MT data performs approximately identically to HT data on the syntactic level. However, the GoogleNMT system performs poorly when dealing with semantic level or issues with additional contextual knowledge, particularly when applying particularization, figuration, equivalence, and generalization with the gap of 57\%, 75\%, 46\%, and 32\% to HT data respectively. When translation relation disparities is analyzed from the perspective of genres, the ues of non-literal translation techniques varies. The usage of non-literal translation techniques between two data sets in official documents, legal texts, and the science sector has a lower gap of 9.8\%, 16.7\%, and 21.5\% respectively, but significantly higher discrepancies exist in microblogs (65\%), scientific papers (81\%), and subtitles (98\%).

In-depth linguistics-related characteristics and variables that influence the usage of non-literal translation techniques are explored in addition to broad comparisons. Elements on semantic and syntactic levels are detected and summarized to investigate 10 non-literal translation techniques. On a semantic level, the translation strategies of equivalence, figurative translation, and generalization are examined. These techniques continue to be broken down into much more granular semantic units. On the other hand, lexical shift, transposition, unaligned explicitation, and reduction need significantly greater syntactic understanding of individual tokens, and the use of automated metrics such POS tagging and dependence relation is the primary way for assessment. Modulation, modulation+transposition, and particularization are further components of a hybrid type that includes both semantic and syntactic aspects. Comparing the two corpora on a linguistic level that results in translation relationships reveals that GoogleNMT performs well at understanding syntactic relationships but poorly at capturing semantic relationships.

\subsection{Shortcoming}

However, there are two inevitable drawbacks in this study. The first problem is that the data distribution of different translation relations is uneven. There are just three aligned pairings for figurative translation, and less than 100 aligned pairs for modulation and modulation + transposition, despite the fact that each genre has 50 phrases. Additionally, the sorting of aligned pairings using semantic knowledge and the annotation of translation relations contain some manual operations, which to some extent adds to the incorrect classification and annotations.

\subsection{Future work}

For future studies and work, there are several things that can be done and fulfilled. In this work, only one machine translation system (Google Translate) is used, and therefore this study lacks an overview of the whole current NMT systems as different NMT systems have their own preferences which lead to different translation outputs. The comparison across several types of NMT systems can render more general and universal conclusions. Besides, an automatic metric to measure the translation quality can be developed based on detecting and recognizing translation relations. Finally, based on the analyzed sub-categories of each translation technique, some directions for NMT system optimization can be provided, and more targeted improvement can be made by integrating certain linguistics-related knowledge into NMT systems.

\pagebreak


\nocite{*}
\bibliographystyle{plain}   

\clearpage
\phantomsection
\addcontentsline{toc}{section}{References}
\tolerance=500
\bibliography{References}

\end{document}